\documentclass{article} 
\usepackage{iclr2027_conference,times}

\usepackage{amsmath,amsfonts,bm}

\def\eqref#1{equation~\ref{#1}}

\def\1{\bm{1}}

\DeclareMathAlphabet{\mathsfit}{\encodingdefault}{\sfdefault}{m}{sl}
\SetMathAlphabet{\mathsfit}{bold}{\encodingdefault}{\sfdefault}{bx}{n}

\usepackage{hyperref}
\usepackage{url}
\usepackage{booktabs}
\usepackage{multirow}
\usepackage{threeparttable}
\usepackage{graphicx}
\usepackage{float}
\usepackage{caption}
\usepackage{tabularx}
\usepackage{makecell}
\usepackage{ragged2e}
\usepackage{algorithm}
\usepackage{algpseudocode}
\usepackage{adjustbox}
\usepackage{array}
\usepackage{xcolor}
\usepackage{tikz} 
\usepackage{lineno}

\usepackage{amsmath,amssymb,amsthm}
\usepackage{tikz}
\usetikzlibrary{arrows.meta,positioning}

\newtheorem{proposition}{Proposition}
\newtheorem{theorem}{Theorem}
\newtheorem{assumption}{Assumption}

\newcommand{\NInstances}{499}           
\newcommand{\NRepos}{12}                
\newcommand{\SolveRate}{93.99\%}        
\newcommand{\ExactRate}{15.23\%}        

\newcommand{\ReplayRate}{1.00}          
\newcommand{\ReplayN}{100}              
\newcommand{\ReplayRepeats}{3}          
\newcommand{\NForkPoints}{10}           
\newcommand{\NForkEpisodes}{3}          
\newcommand{\HighEntFrac}{50\%}         
\newcommand{\EntSpearman}{+0.072}       
\newcommand{\HighEntPos}{0.416}         
\newcommand{\NAttr}{70}                 
\newcommand{\DisagreeRate}{97.1\%}      
\newcommand{\DisagreeCount}{68}
\newcommand{\DisagreeCI}{[92.9\%, 100\%]}
\newcommand{\NDisp}{68}                 
\newcommand{\Displacement}{+0.418}      
\newcommand{\DispZ}{6.07}               
\newcommand{\NColl}{60}                 
\newcommand{\ColliderIso}{+0.537}       
\newcommand{\CollZ}{6.70}
\newcommand{\DegenRate}{22.9\%}         
\newcommand{\DegenCount}{16}
\newcommand{\AUROC}{0.481}
\newcommand{\ECE}{0.015}
\newcommand{\Brier}{0.058}
\newcommand{\AURC}{0.058}
\newcommand{\BaseRate}{93.99\%}
\newcommand{\NCov}{7}                   
\newcommand{\StWCorr}{0.994}            
\newcommand{\StProbMAE}{0.022}          
\newcommand{\StH}{0.464}                
\newcommand{\StUTask}{0.308}
\newcommand{\StUPolicy}{0.146}
\newcommand{\StUModel}{0.012}
\newcommand{\StUModelOOD}{0.100}
\newcommand{\StRatio}{8.1}              
\newcommand{\UPolicy}{0.049}
\newcommand{\NUPolicy}{23}              
\newcommand{\KRep}{4}
\newcommand{\NNondegen}{54}             
\newcommand{\NAttrRepos}{10}            
\newcommand{\NAttrFail}{19}             
\newcommand{\ZeroPointFrac}{54.7\%}     

\newcommand{\CollLaterCount}{59}        
\newcommand{\NEpAll}{491}               
\newcommand{\NEpEval}{488}              
\newcommand{\NPts}{7616}                
\newcommand{\NPtsGraph}{509}            
\newcommand{\FirstVisitShare}{67.0\%}   
\newcommand{\ProvCoverage}{0.689}       
\newcommand{\ProvCoverageCI}{[0.642, 0.739]}
\newcommand{\ProvTopThree}{0.326}       
\newcommand{\ProvTopThreeCI}{[0.276, 0.375]}
\newcommand{\ProvEff}{3.46}             
\newcommand{\GraphTopThree}{0.058}      
\newcommand{\GraphTopThreeCI}{[0.039, 0.078]}
\newcommand{\FreqTopThree}{0.072}       
\newcommand{\ProvVsGraphThree}{+0.136}  
\newcommand{\ProvVsGraphThreeCI}{[+0.102, +0.173]}
\newcommand{\ProvVsGraphTen}{+0.259}
\newcommand{\ProvVsGraphTenCI}{[+0.210, +0.313]}
\newcommand{\MentionedPerEp}{50.5}      
\newcommand{\VisitedPerEp}{6.5}         
\newcommand{\MentionToVisit}{18.0\%}
\newcommand{\RoutedTopOne}{0.150}
\newcommand{\FactoredTopOne}{0.109}
\newcommand{\RoutedGainOne}{+0.041}
\newcommand{\RoutedGainOneCI}{[+0.034, +0.047]}
\newcommand{\RoutedGainTen}{+0.052}
\newcommand{\RoutedGainTenCI}{[+0.047, +0.058]}
\newcommand{\RoutedEff}{12.85}
\newcommand{\FactoredEff}{39.93}
\newcommand{\RoutedECE}{0.012}
\newcommand{\ToolObjMI}{0.159}          
\newcommand{\ToolObjMINull}{0.006}
\newcommand{\ToolObjMIP}{5\times10^{-4}}
\newcommand{\ShareHandler}{27.3\%}
\newcommand{\SharePath}{51.1\%}
\newcommand{\ShareQuery}{21.6\%}
\newcommand{\TopThreeHandler}{0.582}
\newcommand{\TopThreePath}{0.185}
\newcommand{\TopThreeQuery}{0.029}
\newcommand{\AfterSearchThree}{0.469}   
\newcommand{\AfterSearchECE}{0.013}
\newcommand{\AfterReadThree}{0.160}
\newcommand{\QueryDamaged}{75.3\%}      
\newcommand{\ShlexFailRate}{26.6\%}     
\newcommand{\QueryFragDef}{249}          
\newcommand{\QueryTopThreeOld}{0.067}    
\newcommand{\QueryTopThreeRepaired}{0.031}  
\newcommand{\QueryTopThreeBest}{0.076}   
\newcommand{\QueryCovBest}{0.181}
\newcommand{\QueryParseDelta}{-0.036}    
\newcommand{\QueryProblemGain}{+0.041}   
\newcommand{\QuerySymbolGain}{-0.007}    
\newcommand{\NQueryPts}{1661}
\newcommand{\EffectDiffCI}{[-0.008, +0.135]}
\newcommand{\HAbsDeltaRho}{+0.002}       
\newcommand{\RatioAlphaFive}{3.36}
\newcommand{\RatioAlphaTwenty}{1.71}
\newcommand{\OneHopObs}{0.284}          
\newcommand{\OneHopNull}{0.301}
\newcommand{\OneHopLift}{0.94}
\newcommand{\ThreeHopLift}{1.02}
\newcommand{\CommunityLift}{1.01}
\newcommand{\SelfRepeatLift}{1.47}
\newcommand{\ClusterObs}{0.269}         
\newcommand{\ClusterNull}{0.0087}
\newcommand{\ClusterLift}{30.9}
\newcommand{\GoldRankUnseen}{0.049}     
\newcommand{\GoldRankUnseenCI}{[0.021, 0.084]}
\newcommand{\GoldRankPrefix}{0.103}
\newcommand{\GoldVisitedShare}{65\%}    
\newcommand{\GraphResolveRate}{46.2\%}
\newcommand{\NEpReuse}{487}
\newcommand{\ToolReuseRate}{79.1\%}
\newcommand{\ObjReuseRate}{41.6\%}
\newcommand{\ObjSelfRepeat}{1.83}
\newcommand{\ObjSelfRepeatP}{10^{-141}}
\newcommand{\ToolSelfRepeat}{1.02}
\newcommand{\ToolSelfRepeatP}{0.51}
\newcommand{\ToolMILift}{1.13}
\newcommand{\ToolMIP}{10^{-22}}
\newcommand{\GapRatio}{0.90}            
\newcommand{\FreqBias}{1.70}
\newcommand{\FreqBiasCI}{[1.63, 1.76]}
\newcommand{\RecencyBias}{1.74}
\newcommand{\RecencyBiasCI}{[1.67, 1.81]}
\newcommand{\NAttrPrime}{38}            
\newcommand{\NSeg}{190}                 
\newcommand{\MReplay}{8}
\newcommand{\ForkPrime}{1901}           
\newcommand{\HoursPrime}{12.7}
\newcommand{\UexecMean}{0.034}
\newcommand{\UexecZeroShare}{50.5\%}    
\newcommand{\HUexecRho}{-0.088}
\newcommand{\HUexecCI}{[-0.244, +0.062]}
\newcommand{\HUexecInstRho}{-0.219}
\newcommand{\HUexecInstCI}{[-0.518, +0.126]}
\newcommand{\RecallUexecRho}{-0.425}    
\newcommand{\RecallUexecCI}{[-0.695, -0.102]}
\newcommand{\ICCResid}{0.640}           
\newcommand{\ICCResidNull}{0.196}
\newcommand{\ICCResidP}{5\times10^{-4}}
\newcommand{\UexecPosRho}{-0.131}
\newcommand{\UexecToolP}{0.65}
\newcommand{\StepSigRaw}{5.3\%}         
\newcommand{\StepSigBH}{0}              
\newcommand{\JudgeGold}{0.900}
\newcommand{\JudgeGoldCI}{[0.829, 0.971]}
\newcommand{\GoldStepShare}{0.413}      
\newcommand{\JudgeGoldDiff}{+0.487}
\newcommand{\JudgeGoldDiffCI}{[+0.411, +0.557]}
\newcommand{\CausalGold}{0.286}
\newcommand{\FairJudge}{0.826}          
\newcommand{\FairCausal}{0.261}
\newcommand{\FairDiff}{+0.565}
\newcommand{\FairDiffCI}{[+0.304, +0.783]}
\newcommand{\NFair}{23}
\newcommand{\JudgeInEst}{32.9\%}        
\newcommand{\WithinLate}{0.746}         
\newcommand{\WithinLateDiff}{+0.246}
\newcommand{\WithinLateDiffCI}{[+0.172, +0.311]}
\newcommand{\NGoldSteps}{8.7}
\newcommand{\WithinChance}{0.139}
\newcommand{\BestRuleAgree}{0.243}      
\newcommand{\JudgePosMean}{0.751}
\newcommand{\CausalPosMean}{0.333}
\newcommand{\ColliderIsoCI}{[+0.459, +0.610]}
\newcommand{\ColliderMedian}{+0.667}
\newcommand{\ColliderSignP}{1.1\times10^{-16}}
\newcommand{\ColliderTied}{10}          
\newcommand{\PrefixPosMean}{0.214}      
\newcommand{\XModel}{qwen3.6-plus}
\newcommand{\XCalls}{48}                 

\newcommand{\NJudges}{3}
\newcommand{\NCommonEp}{23}
\newcommand{\JudgesPositive}{3}          
\newcommand{\JudgesEnriched}{2}          
\newcommand{\DispCoder}{+0.394}
\newcommand{\DispQwen}{+0.435}
\newcommand{\DispOrig}{+0.534}
\newcommand{\DispSpread}{0.14}           
\newcommand{\HetDispP}{0.181}
\newcommand{\HetEnrichP}{0.044}
\newcommand{\EnrichCoder}{0.826}
\newcommand{\EnrichQwen}{0.565}
\newcommand{\EnrichOrigCommon}{0.826}
\newcommand{\EnrichBaseCommon}{0.443}

\newcommand{\NoCompaction}{0}           
\newcommand{\RecallSens}{2.505}         
\newcommand{\PctMoved}{98.6\%}          
\newcommand{\PctMovedY}{73.2\%}         
\newcommand{\GoldProvEst}{0.362}        
\newcommand{\FOneSens}{2.165}
\newcommand{\YSens}{1.643}

\title{What Process Evaluation of Coding Agents Actually Measures: Action, Task, and Step Are Three Different Levels}

\author{
Jiawei He$^{1,\dagger}$, Mengyu Shi$^{2,\dagger}$, Jie jia$^{1}$,\\
Xikai Yang$^{1,*}$, Dong Sun$^{1,*}$, \\[0.5em]
\normalsize
$^{1}$Amap, Alibaba Group, China\\
$^{2}$State Key Laboratory for Novel Software Technology, Nanjing University, Nanjing, China\\
$^{\dagger}$Contribute equally \quad
$^{*}$Corresponding author
}

\iclrfinalcopy 
\begin{document}

\maketitle
\begin{abstract}
Coding agents are increasingly evaluated not only by whether they solve a task, but also by how they execute it. However, existing process-level evaluations often treat action prediction, task uncertainty, and step attribution as if they were the same problem, which makes it unclear what such evaluations actually measure. In this paper, we introduce a measurement framework for process evaluation in coding agents and instantiate step-level causal attribution with SCAE, a replay-based estimator derived from a structural causal model of agent execution. Our framework combines prefix-conditioned identification, replay/intervention-based estimation, and controlled judge-information manipulation to study process evaluation at the action, task, and step levels. Experiments on \NInstances{} file-localization episodes from \NRepos{} repositories show that next actions are driven primarily by execution provenance rather than code-graph transitions, execution uncertainty is structured at the task rather than step level, and full-trace judges exhibit systematic collider bias, suggesting that current process evaluation often measures semantic relevance rather than certified causal contribution.
\end{abstract}
\section{Introduction}
\label{sec:intro}

Large language model (LLM) coding agents are increasingly evaluated not only by whether they solve a task, but also by how they execute it~\citep{si2026ideation}. In practice, developers want more than a final success label: they want to understand which steps advanced the task, which steps caused failure, whether the execution remained interpretable and controllable, and whether process-level signals can be trusted for debugging, benchmarking, or training~\citep{prasai2026knowthyself}. As coding-agent trajectories grow longer and more tool-centric, these questions become increasingly difficult to answer from final outcomes alone~\citep{he2026procctrlbench}.

Despite its growing importance, current process evaluation remains conceptually underspecified. Existing methods often assume that a trajectory can be meaningfully evaluated step by step, but in practice they mix together three different levels of analysis~\citep{wang2025steca,sun2026llm}. At the action level, the question is what the agent is likely to do next~\citep{wen2024reinforcing}. At the task level, the question is whether the overall execution remains uncertain or recoverable~\citep{wang2025megaagent}. At the step level, the question is whether a particular action causally changed the final outcome~\citep{wang2025steca}. Process reward models~\citep{setlur2025rewarding}, full-trace LLM judges~\citep{chen2026seeing}, and restart-based ablations~\citep{kontonis2026memento} are frequently discussed as if they answered the same question, even though they condition on different information and target different quantities. This makes their outputs difficult to interpret and even harder to validate.

In this paper, we argue that action, task, and step are three different levels, and that process evaluation should distinguish them explicitly. To study this problem, we introduce a measurement framework for coding-agent process evaluation and instantiate step-level attribution with SCAE, a replay-based estimator derived from a structural causal model of agent execution. Rather than presenting SCAE primarily as a better attribution tool, we use it as a measurement instrument: by making step-local causal effects explicit and estimable in principle, it allows us to test what process evaluation can and cannot identify in practice.

Our setting is file localization, where an episode consists of an issue, a repository snapshot, a sequence of agent actions, and a verifiable target set of files. This setting is especially useful because it provides the ingredients required for principled process analysis: replayable environments, fully logged prefixes, and a checkable terminal outcome. We reconstruct prefixes, replay continuations, and estimate step-local effects under explicit assumptions about context observability, decoding noise, and environment replayability. When local overlap is available, the effect is estimated observationally; when the policy is effectively deterministic, we switch to an intervention branch. The same framework also allows us to analyze next-action prediction, uncertainty structure, and attribution behavior within a unified experimental design.

Our experiments reveal a consistent picture. At the action level, the next operation is often predictable, but not from the structure one might expect: the strongest signal is the agent's own execution provenance, namely the paths it has just seen in tool outputs, rather than the code dependency graph. At the task level, execution variance is structured mainly by the instance rather than by individual steps, which implies that uncertainty is primarily a property of the task. At the step level, the identifiable causal quantity proves too weak to localize at feasible replay cost, yet full-trace judges still shift their blame systematically when later steps are made visible. This isolates a bias in the evaluation mechanism itself and suggests that many current process-level evaluators measure semantic relevance or late-trace evidence rather than certified causal contribution.

These findings lead to a simple conclusion: process evaluation is not one problem but several. Some process-level signals are useful because they predict what the agent will do next. Others are useful because they characterize how uncertain the task remains. But these signals should not automatically be interpreted as step-level causal attribution. Once the levels are separated, the apparent tractability of process evaluation becomes easier to understand: action prediction can be strong, task-level uncertainty can be structured, and step-level causal localization can still remain infeasible.

Our main contributions are as follows:
\begin{itemize}
    \item We formulate process evaluation in coding agents as a three-level measurement problem, distinguishing action prediction, task uncertainty, and step attribution instead of treating them as a single notion of process quality.
    \item We introduce SCAE, a replay-based measurement framework grounded in a structural causal model, and use it to study what step-level process evaluation can identify in principle and what it can measure in practice.
    \item We show empirically that next actions are governed primarily by execution provenance rather than code-graph transitions, and that execution uncertainty resides mainly at the task rather than step level.
    \item We isolate collider bias in full-trace LLM judging through a controlled manipulation of the judge's information set, showing that current process attribution is better interpreted as semantic relevance than as certified causal contribution.
\end{itemize}

\section{Related Work}
\label{sec:related_work}

\paragraph{Counterfactual Reasoning in Discrete Sequential Systems.}
Counterfactual reasoning over discrete action sequences is commonly grounded in the Gumbel--max structural causal model~\citep{ravfogel2025gumbel}, which also provides the core mechanism underlying SCAE. 
By representing categorical sampling as perturb-and-argmax~\citep{khanmohammadi2025calibrating}, this line of work yields a tractable posterior over latent noise variables conditioned on an observed draw. 
Building on this formulation, prior studies~\citep{chatzi2025counterfactual,kazemi2024counterfactual} have developed counterfactual explanations for action sequences under known MDPs, extended the framework to token-level generation, and analyzed how counterfactual influence decays as trajectories diverge. 
However, these approaches generally assume access to known transition mechanisms and a fixed action space. 
In contrast, LLM agents operating over code repositories provide neither. 
This introduces two challenges that are largely absent in standard simulators: positivity may fail locally at individual decision points, and context compaction may destroy the invertibility of the state mapping. Accordingly, SCAE does not assume a known transition mechanism; instead, it estimates one through replay. It also treats positivity as an empirical property to be measured rather than a standing assumption, and regards context compaction as an identification failure rather than an approximation error.

\paragraph{Step-Level Credit Assignment and Process Supervision.}
Process reward models supervise intermediate reasoning steps and have been shown to outperform outcome-only supervision on reasoning tasks \citep{zhang2026linking,setlur2025rewarding,wang2024mathshepherd}. 
Their primary purpose, however, is to improve policy learning rather than to explain the role of a particular step in a single trajectory. 
In practice, their supervision signals are often derived from Monte Carlo estimates over completed rollouts, for example by judging whether a given prefix can still reach a correct answer. 
As a result, information from downstream steps is reintroduced into the label, and the contribution of the current step is not cleanly separated from the effect of what follows. 
A related observation applies to hindsight and counterfactual credit assignment, which deliberately use future information to reduce the variance of policy-gradient estimates \citep{arjonamedina2019rudder,liang2026rlhs,yang2026int}. 
Such quantities are well suited for optimization, but they do not correspond to per-episode causal contributions. 
Another relevant line of work concerns axiomatic attribution methods, such as Shapley values \citep{horovicz2024tokenshap}, which satisfy an additive property analogous to our telescoping identity, but at the cost of exponentially many re-evaluations. 
In contrast, our goal is not to propose a stronger step scorer. 
Rather, we clarify what information a step score must condition on in order to admit a causal interpretation, show that this distinction systematically changes which step is blamed, and provide an estimator that preserves additivity through a single fitted object instead of requiring $2^T$ re-executions.

\paragraph{Process-Oriented Evaluation of Coding Agents.}
Current evaluation of coding agents remains centered primarily on terminal task success \citep{merrill2026terminal,xu2026theagentcompany,xia2024agentless}. 
In parallel, a complementary line of work studies recurrent execution defects and calibrates detector outputs against human process annotations \citep{he2026procctrlbench}. 
These approaches target a different estimand: the probability that an annotator would mark a step as defective given trajectory evidence~\citep{zhang2026agentracer,li2026trajectories}. 
As such, they address a measurement problem about annotation agreement rather than a causal inference problem. By contrast, our focus is the causal effect of substituting one step on the final outcome. This estimand requires that the randomization induced by the logging policy be preserved and is not identifiable under context compaction (\S\ref{sec:identification}). Because the two estimands are fundamentally different, a step may be consistently judged defective by annotators while contributing no identifiable causal effect to the final outcome. Annotation agreement alone therefore cannot establish whether a process score measures what it is intended to measure. SCAE differs in that it evaluates attribution against a verifiable terminal outcome, thereby turning the question of what a process score measures into an empirically testable one rather than a purely definitional one.

\paragraph{Uncertainty Quantification for Language Models.}
Research on calibration and confidence estimation for language models has largely focused on single-turn generation, where each output is associated with a single scalar confidence value \citep{xia2025influences,ulmer2024calibrating,nakkiran2026trained}. More broadly, decomposing predictive uncertainty into aleatoric and epistemic components is standard in Bayesian deep learning \citep{melo2024deep,ross2026textual}, while conformal prediction provides distribution-free coverage guarantees under exchangeability \citep{vovk2005algorithmic,angelopoulos2023conformal}. These tools are well suited to settings in which one prediction corresponds to one confidence estimate. The agent setting introduces two complications that they do not directly address. First, uncertainty compounds across sequential decisions. Second, the agent's own actions determine what information becomes observable later, making uncertainty inherently trajectory-dependent. 
Consequently, a single terminal confidence score cannot reveal where reliability was lost, nor does it specify the level of the trajectory at which confidence should be attached. File localization itself is a classical subproblem in automated program repair, typically evaluated by retrieval-style accuracy against the files touched by a reference patch \citep{wang2026improving,xia2024agentless}. We adopt localization not as a performance target to improve, but as the minimal controllable setting in which the identification conditions required by our analysis can be satisfied. On this basis, we treat the level at which uncertainty resides as an empirical question rather than a modeling choice, and examine whether uncertainty is primarily a property of the task or of individual steps.

\paragraph{Structural Priors for Agent Navigation.}
Existing coding-agent systems make substantial use of repository structure. Retrieval and localization modules commonly rank files according to similarity or dependency information, and agent scaffolds expose repository-shaped tools through which the model navigates the codebase \citep{xu2026gcig,,edge2024local}. As a result, the assumption that code structure guides the agent's next move is already built into current system designs. Empirically, the files visited within a single episode do tend to cluster in the dependency graph. However, such clustering does not by itself imply that local step-to-step transitions are governed by graph structure. To our knowledge, prior work has not explicitly tested this assumption against a null model that holds the working region fixed, and therefore has not separated descriptive clustering from the actual transition mechanism. Our analysis fills this gap and shows that repository structure is better understood as a prior over the agent's working region than as a local transition model for next-step prediction. Moreover, we identify an alternative structural signal---the agent's own execution provenance---that is more directly predictive of local behavior. This signal provides a more appropriate conditioning basis for step-level prediction and context-selection policies.

\section{Method}
\label{sec:method}

This section presents our methodological framework for measuring process evaluation at the action, task, and step levels. Our goal is not only to define a step-level causal quantity, but also to build a practical measurement pipeline that can be instantiated on real coding-agent traces. To this end, we introduce SCAE, a replay-based framework grounded in a structural causal model of agent execution. SCAE serves two roles in this paper. First, it provides the formal object needed to distinguish step-local causal attribution from other process-level signals. Second, it acts as a measurement instrument for testing what process evaluation can and cannot identify in practice.

At a high level, the framework takes as input a coding-agent trajectory, the corresponding repository snapshot, and a verifiable task outcome, and produces three kinds of measurements. At the action level, it characterizes what structure predicts the next operation. At the task level, it measures execution uncertainty under replayed continuations. At the step level, it estimates prefix-conditioned causal effects when the required assumptions hold. The method proceeds in three stages. We first formalize agent execution with a structural causal model and specify the conditions under which step-local effects are identified. We then estimate an anytime value function by replaying reconstructed prefixes, switching between observational replay and intervention depending on local overlap. Finally, we derive attribution, forecasting, and uncertainty quantities from the same fitted object and use them in the experiments of Section~\ref{sec:experiments}.

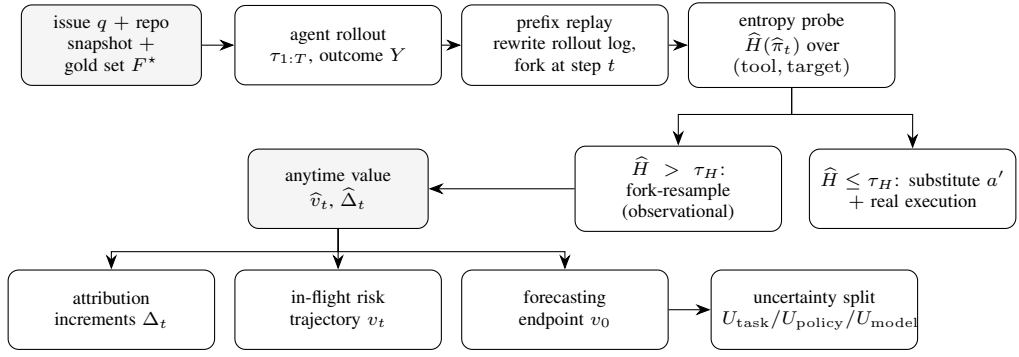
\begin{figure}[t]
\centering
\begin{tikzpicture}[
  font=\small,
  stage/.style={draw,rounded corners,align=center,inner sep=4pt,
                minimum height=1.05cm,text width=2.45cm,font=\scriptsize},
  data/.style={draw,fill=black!4,rounded corners,align=center,inner sep=4pt,
               minimum height=1.05cm,text width=2.1cm,font=\scriptsize},
  ar/.style={-{Stealth[length=2mm]}},
  note/.style={font=\scriptsize,align=center}
]
\node[data]  (ep)   at (0,0)    {issue $q$ $+$ repo snapshot $+$ gold set $F^\star$};
\node[stage] (roll)  at (3.0,0)  {agent rollout\\ $\tau_{1:T}$, outcome $Y$};
\node[stage] (fork)  at (6.0,0)  {prefix replay\\ rewrite rollout log, fork at step $t$};
\node[stage] (ent)   at (9.0,0)  {entropy probe\\ $\widehat H(\widehat\pi_t)$ over\\ $(\mathrm{tool},\mathrm{target})$};

\draw[ar] (ep) -- (roll);
\draw[ar] (roll) -- (fork);
\draw[ar] (fork) -- (ent);

\node[stage] (obs) at (7.5,-1.9)
  {$\widehat H > \tau_H$: fork-resample\\ (observational)};
\node[stage] (int) at (10.6,-1.9)
  {$\widehat H \le \tau_H$: substitute $a'$\\ $+$ real execution};
\draw[ar] (ent.south) -- ++(0,-0.35) -| (obs.north);
\draw[ar] (ent.south) -- ++(0,-0.35) -| (int.north);

\node[data] (v) at (3.0,-1.9) {anytime value\\ $\widehat v_t$, $\widehat\Delta_t$};
\draw[ar] (obs.west) -- (v.east);

\node[stage] (attr) at (0.0,-3.5) {attribution\\ increments $\Delta_t$};
\node[stage] (risk) at (3.0,-3.5) {in-flight risk\\ trajectory $v_t$};
\node[stage] (pred) at (6.0,-3.5) {forecasting\\ endpoint $v_0$};
\draw[ar] (v.south) -- ++(0,-0.3) -| (attr.north);
\draw[ar] (v.south) -- (risk.north);
\draw[ar] (v.south) -- ++(0,-0.3) -| (pred.north);

\node[stage] (uq) at (9.3,-3.5)
  {uncertainty split\\ $U_{\mathrm{task}}/U_{\mathrm{policy}}/U_{\mathrm{model}}$};
\draw[ar] (pred.east) -- (uq.west);
\end{tikzpicture}
\caption{\textbf{Overview of SCAE}. An episode is rolled out once to obtain
$(\tau_{1:T}, Y)$. Prefix replay reconstructs $\tau_{1:t}$ by rewriting the
rollout log at item granularity, which requires no modification to the agent.
An entropy probe at each estimation point decides which identification strategy
applies: where the policy retains randomness, $v_t$ is estimated observationally
by fork-resampling (Proposition~\ref{prop:seqrand}); where it is effectively
deterministic, an alternative action is substituted and \emph{actually executed}
(Assumption~\ref{as:a5}). The single fitted object $\widehat v_t$ then supports
three readings. We report what each delivered rather than what it promised: the
increments do not localize at feasible cost (\S\ref{sec:rq4}), the trajectory
yields a task-level rather than step-level uncertainty
(\S\ref{sec:rq4}), and the endpoint $v_0$ does not forecast the outcome
(leave-one-repository-out AUROC \AUROC{}). The decomposition of
\S\ref{sec:uq} is validated on a synthetic generating process, not on these
traces. In the figure, shaded boxes are data and plain boxes are pipeline stages;
the two branches below the entropy probe are the two cases of \eqref{eq:hybrid}.
Algorithm~\ref{alg:scae} in Appendix~\ref{sec:algorithm} states the procedure.}
\label{fig:pipeline}
\end{figure}

\subsection{Method overview}
\label{sec:method_overview}

Figure~\ref{fig:pipeline} summarizes the overall pipeline. Starting from an issue, a repository snapshot, and a gold target set, we execute the agent once to obtain a factual trajectory and a terminal outcome. We then reconstruct selected prefixes from the recorded trace and resume execution from those prefixes multiple times. These replays provide two kinds of information. First, they estimate an anytime value function that tracks how the expected terminal outcome changes over the course of the trajectory. Second, they probe whether the local action distribution retains enough randomness for observational identification to be meaningful.

This distinction is important because overlap is not globally available in coding-agent execution. Some decision points remain stochastic under replay, while others are effectively deterministic. SCAE therefore uses a hybrid estimation strategy: if replayed continuations reveal local action variation, step-level quantities are estimated observationally from replay; if the policy is effectively deterministic, we substitute an alternative action and execute the resulting continuation in the real environment. This hybrid design is not introduced for convenience but forced by the data-generating process.

The same estimated object supports three downstream readings. The increments of the anytime value define step-level attribution quantities. The left endpoint defines a pre-execution forecast. The replay distribution around each point defines execution uncertainty and its decomposition. In this way, one framework supports all three levels studied in the paper while making explicit that they are not interchangeable.

\subsection{Problem formulation}
\label{sec:problem_formulation}

We study step-level process evaluation through a prefix-conditioned causal quantity. Let $\tau_{1:t-1}$ denote the realized execution prefix before step $t$, let $A_t$ denote the action taken at that step, and let $Y$ denote the terminal task outcome. The step-level effect of replacing action $a$ with an alternative action $a'$ is defined as
\begin{equation}
\theta_t(a, a')
\;=\;
\mathbb{E}\big[Y_{A_t = a'} - Y_{A_t = a} \mid \tau_{1:t-1}\big].
\label{eq:estimand}
\end{equation}
This estimand asks whether changing a single action, while holding the realized history fixed, would change the final outcome.

This quantity is different from the targets implicitly measured by several common process-evaluation practices. Full-trace judges condition on downstream actions and the final outcome; restart-based ablations discard the realized history and resample the whole continuation; process reward labels are often derived from completed trajectories. These are not only procedural differences. They correspond to different conditioning sets, and therefore to different measurement targets. Our framework is designed to make this distinction explicit and to test what can be identified under the prefix-conditioned target in Eq.~\eqref{eq:estimand}; Appendix~\ref{sec:formal-app} tabulates each practice, the bias it incurs, and the empirical signature it should leave.

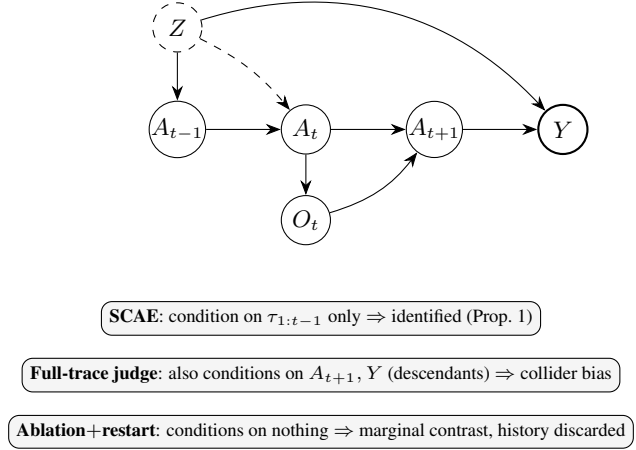
\begin{figure}[t]
\centering
\begin{tikzpicture}[
  node distance=1.05cm and 1.25cm,
  var/.style={circle,draw,minimum size=6.5mm,inner sep=0pt,font=\small},
  lat/.style={circle,draw,dashed,minimum size=6.5mm,inner sep=0pt,font=\small},
  outnode/.style={circle,draw,thick,minimum size=6.5mm,inner sep=0pt,font=\small},
  ar/.style={-{Stealth[length=2mm]}},
  lbl/.style={font=\scriptsize},
  box/.style={draw,rounded corners,inner sep=3pt,font=\scriptsize}
]
\node[lat] (Z)  at (0,1.35) {$Z$};
\node[var] (Ap) at (0,0)    {$A_{t-1}$};
\node[var] (At) at (1.7,0)  {$A_t$};
\node[var] (An) at (3.4,0)  {$A_{t+1}$};
\node[outnode] (Y)  at (5.1,0)  {$Y$};
\node[var] (Ot) at (1.7,-1.2) {$O_t$};

\draw[ar] (Ap) -- (At);
\draw[ar] (At) -- (An);
\draw[ar] (An) -- (Y);
\draw[ar] (At) -- (Ot);
\draw[ar] (Ot) to[bend right=18] (An);
\draw[ar] (Z) -- (Ap);
\draw[ar] (Z) to[bend left=32] (Y);
\draw[ar,dashed] (Z) to[bend left=12] (At);

\node[box,fill=black!4] (c1) at (1.9,-2.5)
  {\textbf{SCAE}: condition on $\tau_{1:t-1}$ only $\Rightarrow$ identified (Prop.~\ref{prop:seqrand})};
\node[box,fill=black!4] (c2) at (1.9,-3.25)
  {\textbf{Full-trace judge}: also conditions on $A_{t+1},Y$ (descendants) $\Rightarrow$ collider bias};
\node[box,fill=black!4] (c3) at (1.9,-4.0)
  {\textbf{Ablation$+$restart}: conditions on nothing $\Rightarrow$ marginal contrast, history discarded};
\end{tikzpicture}
\caption{\textbf{Three conditioning sets and their causal status}. The
conditioning set, not the scorer, determines whether attribution is causal. Latent
task difficulty $Z$ (dashed) influences both behaviour and outcome, but
conditioning on the prefix $\tau_{1:t-1}$ blocks it, because the policy observes
nothing that is not logged and decoding noise is exogenous. Conditioning
\emph{additionally} on $A_{t+1}$ and $Y$, which is what full-trace judging does,
conditions on descendants of $A_t$ and opens a biasing path. The predicted
signature is attribution displaced toward later steps. We test it in
\S\ref{sec:rq5} by a controlled manipulation of the judge's information set,
which is the one attribution result here that requires no causal reference.}
\label{fig:teaser}
\end{figure}

Figure~\ref{fig:teaser} illustrates the core causal distinction. The valid conditioning set for step-local attribution is the realized prefix only. Conditioning additionally on later actions or on the terminal outcome changes the estimand and, in the case of full-trace judging, introduces post-treatment bias. This is why process evaluation can appear informative while still failing to provide valid step-level causal attribution. A related conflation, in which data-flow reachability is offered as a proxy for causal irrelevance, comes apart in both directions; Appendix~\ref{sec:dataflow} gives counterexamples.

\subsection{An agent--environment structural causal model}
\label{sec:model}

Let $\mathcal{M} = \langle U, V, F, P(U)\rangle$ be a structural causal model with exogenous variables
$$
U = \big(Z, \{G^{(t)}_\pi\}, \{U^{(t)}_{\mathrm{env}}\}\big),
$$
where $Z$ denotes latent task difficulty, $G^{(t)}_\pi$ denotes decoding noise, and $U^{(t)}_{\mathrm{env}}$ denotes environment noise. The endogenous system is defined by
\begin{align}
A_t &:= \arg\max_k \big(\log \pi_k(S_t) + G^{(t)}_{\pi,k}\big),
   \qquad G^{(t)}_{\pi,k} \sim \mathrm{Gumbel}(0,1),
   \label{eq:action}\\
O_t &:= \mathcal{E}\big(A_t, G;\, U^{(t)}_{\mathrm{env}}\big),\\
S_{t+1} &:= \mathcal{T}(S_t, A_t, O_t), \qquad S_1 := \sigma(q, \mathrm{sys}),\\
\hat F &:= \phi(S_T), \qquad Y := \mathbb{1}\big[F^\star \subseteq \hat F\big].
\end{align}

Two modeling choices are central. First, the policy is written in Gumbel--max form, which makes categorical decisions representable as perturb-and-argmax draws and gives a tractable account of local policy stochasticity. Second, the task outcome is defined by containment rather than exact set equality, because localization should not be penalized simply for naming reasonable auxiliary files such as tests. This outcome remains verifiable while aligning better with the semantics of repository search.

\subsection{Identification}
\label{sec:identification}

We now state the conditions under which the prefix-conditioned effect in Eq.~\eqref{eq:estimand} is identified.

\begin{assumption}[Context observability]
\label{as:a1}
$S_t$ is a deterministic function of $(q, \tau_{1:t-1})$.
\end{assumption}

\begin{assumption}[Exogenous decoding noise]
\label{as:a2}
$G^{(t)}_\pi \perp (Z, U^{(<t)}, U_{\mathrm{env}})$.
\end{assumption}

\begin{assumption}[Replayable environment]
\label{as:a3}
Given $(A_t, G, U^{(t)}_{\mathrm{env}})$, the observation $O_t$ is determined.
\end{assumption}

\begin{proposition}[Sequential randomization by construction]
\label{prop:seqrand}
Under Assumptions~\ref{as:a1}--\ref{as:a3}, we have
$A_t \perp (Z, U^{(\ge t)}) \mid \tau_{1:t-1}$. Consequently, for any action $a$
with $\pi(a \mid \tau_{1:t-1}) > 0$,
\begin{equation}
\mathbb{E}\big[Y \mid \tau_{1:t-1}, do(A_t{=}a)\big]
=
\mathbb{E}\big[Y \mid \tau_{1:t-1}, A_t{=}a\big].
\label{eq:ident}
\end{equation}
Thus, step-level causal effects are nonparametrically identified from observational traces given the realized prefix.
\end{proposition}

The intuition is straightforward. Conditioning on the realized prefix fixes the state $S_t$, so the remaining variation in $A_t$ comes only from exogenous policy noise. Since this noise is independent of latent task difficulty and of future continuation noise, the step-$t$ action behaves as a locally randomized treatment assignment. This yields identification of the prefix-conditioned causal effect without the usual no-unmeasured-confounding assumption. Appendix~\ref{sec:proofs} gives the full argument and shows exactly which step of it fails under each departure from this conditioning set.

The assumptions are also empirically meaningful. Context observability fails under compaction, because dropped context becomes unobserved state. Replayability fails if the environment cannot be reproduced from the logged state and action. Decoding exogeneity is a structural property of the serving policy rather than something directly testable from traces. We therefore verify the first two assumptions empirically in the experimental setup and treat the third as a modeling assumption. Appendix~\ref{sec:breaks} states the three failure modes formally, none of which is confounding.

\subsection{An anytime value and step-level contribution}
\label{sec:estimator}

Rather than estimating each step effect independently, we define an anytime value function along the trajectory:
\begin{equation}
v_t := \mathbb{E}[Y \mid \tau_{1:t}, q, G],
\end{equation}
and its increment
\begin{equation}
\Delta_t := v_t - v_{t-1}.
\end{equation}
This construction supports a natural attribution view in which each step contributes by changing the expected terminal outcome relative to the immediately preceding prefix.

\begin{theorem}[Exactly additive step contributions]
\label{thm:additive}
Under Assumptions~\ref{as:a1}--\ref{as:a3},
\begin{equation}
\sum_{t=1}^{T} \Delta_t = v_T - v_0 = Y - \mathbb{E}[Y \mid q, G].
\label{eq:telescope}
\end{equation}
Moreover, each increment can be decomposed into an action effect relative to the policy's own baseline and an observation effect induced by the returned tool output.
\end{theorem}

This theorem shows that the total surprise of an episode, relative to its pre-execution expectation, can be distributed over steps without residual. In practice, the quantity of interest is not only whether the final outcome changed, but when and where the expected outcome changed along the trajectory. The increments $\Delta_t$ therefore serve as our step-level attribution object. Appendix~\ref{sec:proofs} proves the theorem and states the action/observation split explicitly, since the two halves imply different remedies: better planning versus better tools or indexing.

\subsection{Replay-based estimation under local positivity}
\label{sec:hybrid}

The quantities above are population-level objects. To estimate them on real coding-agent traces, we reconstruct a factual prefix, resume execution from that prefix multiple times, and average the resulting outcomes. Concretely, for an estimation point $t$, we replay the continuation from $\tau_{1:t}$ for $m$ repetitions and estimate
\begin{equation}
\widehat v_t
=
\frac{1}{m}\sum_{i=1}^{m} Y^{(i)}_t,
\end{equation}
where $Y^{(i)}_t$ is the terminal outcome of the $i$-th replay from prefix $\tau_{1:t}$. The corresponding empirical contribution is then
\begin{equation}
\widehat \Delta_t = \widehat v_t - \widehat v_{t-1}.
\end{equation}
Because all arms share the reconstructed prefix and the pinned environment noise, this is a common-random-numbers design, which is what makes a small replay budget usable at all.

This replay estimator is valid only when the local action distribution retains sufficient overlap. In coding-agent execution, however, positivity is not globally available: some prefixes generate diverse next actions under replay, while others repeatedly produce the same action and therefore provide no observational support for nearby alternatives. We therefore treat positivity as a local property that must be measured rather than assumed.

To detect whether local overlap is present, we probe the replayed action distribution at each estimation point. Let $\mathcal{C}$ denote action buckets defined at the level of tool and target object. Using $m$ replayed actions at step $t$, we define the empirical action distribution
\begin{equation}
\widehat{\pi}_t(c)
=
\frac{1}{m}\sum_{i=1}^{m}\mathbb{1}\big[\mathrm{bucket}(A_t^{(i)})=c\big],
\qquad c\in\mathcal{C},
\end{equation}
and its empirical entropy
\begin{equation}
\widehat H(\widehat \pi_t)
=
-\sum_{c\in\mathcal{C}}\widehat{\pi}_t(c)\log \widehat{\pi}_t(c).
\label{eq:entropy}
\end{equation}
Intuitively, $\widehat H(\widehat \pi_t)$ measures whether the next action remains stochastic under replay from the same realized prefix. Appendix~\ref{sec:entropy-app} reports the measured entropy at every probed decision point, which is what shows that overlap varies \emph{within} an episode rather than globally.

This leads to a hybrid estimation strategy. When the replayed action entropy is positive, we use observational replay and estimate the contribution by the difference of replay-based value estimates. When the entropy is effectively zero, we switch to an intervention branch that substitutes an alternative action and executes the resulting continuation in the real environment. Formally,
\begin{equation}
\resizebox{0.92\linewidth}{!}{$
\widehat{\Delta}_t =
\begin{cases}
\widehat v_t - \widehat v_{t-1},
& \text{if } \widehat H(\widehat \pi_t) > \tau_H
\quad \text{(observational branch)},\\[6pt]
\widehat{\mathbb{E}}\!\left[Y \mid \tau_{1:t-1}, do(A_t=a')\right]
-
\widehat{\mathbb{E}}\!\left[Y \mid \tau_{1:t-1}, do(A_t=a_t)\right],
& \text{if } \widehat H(\widehat \pi_t) \le \tau_H
\quad \text{(intervention branch)}.
\end{cases}
$}
\label{eq:hybrid}
\end{equation}

The intervention branch is needed because a deterministic policy step provides no observational support for nearby counterfactual alternatives. In that case, we substitute an alternative action $a'$ into the reconstructed prefix, execute it for real, splice its genuine tool output into the continuation, and then replay the rest of the trajectory. This preserves the semantics of an action intervention. We never fabricate tool outputs, because replacing an observation rather than an action would target a different causal quantity.

\begin{assumption}[Perturbation transferability]
\label{as:a5}
The substituted action $a'$ lies in the support of a suitably perturbed version of the deployed policy.
\end{assumption}

Assumption~\ref{as:a5} makes explicit that the low-entropy branch is not observational identification under the original policy. Instead, it estimates the effect of moving from the realized action to a nearby action that remains behaviorally plausible under a perturbed policy. We label this distinction explicitly in the experiments, and Appendix~\ref{sec:ladder-position} states where the resulting estimates sit on the causal ladder.

\subsection{Outcome choice and practical estimation}
\label{sec:practical_estimation}

The framework above applies to any bounded terminal outcome. In principle, one could use a binary success variable, exact file-set equality, or recall over the gold file set. In practice, this choice matters because it determines how much signal remains available to replay-based estimation. Binary success is often too coarse in file localization: once a trajectory already fails, many step substitutions do not change the indicator at all. Exact set equality is even more brittle because it penalizes otherwise reasonable auxiliary predictions such as test files. We therefore use recall-based outcomes in the main experiments, which preserve more variation and make step- and task-level changes empirically detectable. Appendix~\ref{sec:outcome-app} reports the screening of all six candidate outcomes and the degenerate rate that motivated the choice.

Replay-based estimation also requires selecting a finite set of estimation points and a finite replay budget per point. We therefore estimate values at a set of sampled prefixes rather than at every step in the trajectory. The resulting contributions should be interpreted as segment-level approximations rather than per-token or per-tool-call infinitesimals. This design trades off granularity against replay cost and is one reason we later distinguish what is formally identifiable from what is empirically measurable at feasible scale.

\subsection{Uncertainty decomposition}
\label{sec:uq}

The same replay-based value function also supports uncertainty analysis. Let
\begin{equation}
v_0 = \mathbb{E}[Y \mid q, G]
\end{equation}
denote the pre-execution success probability conditioned only on the task description and repository structure. Its predictive uncertainty can be decomposed conceptually into three components:
\begin{equation}
\mathbb{H}[Y \mid q, G]
=
U_{\mathrm{task}}
+
U_{\mathrm{policy}}
+
U_{\mathrm{model}}.
\label{eq:decomp3}
\end{equation}
Here $U_{\mathrm{task}}$ denotes irreducible uncertainty arising from task ambiguity, $U_{\mathrm{policy}}$ denotes variability induced by stochastic agent execution, and $U_{\mathrm{model}}$ denotes epistemic uncertainty in the forecasting model itself.

This decomposition is useful because the three terms correspond to different operational responses. High task uncertainty suggests ambiguity in the problem itself; high policy uncertainty suggests repeated execution and aggregation; high model uncertainty suggests abstention or escalation. In this paper, however, we do not claim that all three components are directly and fully identified on real traces. Our empirical use of this decomposition is more modest. We estimate policy-driven execution variance directly from replay, use a simple forecasting model to probe the left endpoint $v_0$, and test whether the uncertainty relevant to execution outcome is structured at the step level or at the task level. The decomposition therefore serves as a principled organizational framework for uncertainty analysis rather than as a fully solved inference problem in the main benchmark setting. Appendix~\ref{sec:uq-table} gives an estimator for each component and the synthetic self-test on which the three-way separation is validated; Appendix~\ref{sec:v0-app} gives the $v_0$ model and Appendix~\ref{sec:conformal} the set-valued variant.

\subsection{Evaluation quantities}
\label{sec:metrics}

The experiments evaluate process-level measurements rather than only agent outcomes, so we define the main reported quantities here.

For attribution comparisons, let $\hat t_i^M$ denote the step selected by method $M$ in episode $i$, and let $\mathrm{pos}(t)=t/T_i$ denote normalized position within the trajectory. We compare two attribution methods using top-1 disagreement and positional displacement:
\begin{equation}
D_1(M,M')
=
\frac{1}{N}\sum_{i=1}^{N}
\mathbb{1}\big[\hat t_i^M \neq \hat t_i^{M'}\big],
\qquad
D_3(M,M')
=
\frac{1}{N}\sum_{i=1}^{N}
\big(\mathrm{pos}(\hat t_i^M)-\mathrm{pos}(\hat t_i^{M'})\big).
\label{eq:d1d3}
\end{equation}
In this paper, the displacement statistic is more informative than raw disagreement, because disagreement against an uncertified causal reference is not itself evidence of bias, whereas systematic displacement remains meaningful even when the reference is noisy. Appendix~\ref{sec:metrics-app} works out why $D_1$ is near its ceiling under the null and should therefore be retired.

When a method selects a blamed step, we also ask whether that step aligns with the most harmful estimated segment, whether it lands on a zero-contribution segment, and what signed contribution it receives. Let $\sigma_i(t)$ denote the estimation segment containing step $t$, and let $\mathcal N$ be the subset of non-degenerate episodes whose estimated contributions are not identically zero. We define
\begin{equation}
R(M)
=
\frac{1}{|\mathcal N|}\sum_{i\in\mathcal N}
\mathbb{1}\!\left[\sigma_i(\hat t_i^M)=\arg\min_t \widehat{\Delta}_{i,t}\right],
\end{equation}
\begin{equation}
Z(M)
=
\frac{1}{|\mathcal N|}\sum_{i\in\mathcal N}
\mathbb{1}\!\left[\widehat{\Delta}_{i,\sigma_i(\hat t_i^M)}=0\right],
\end{equation}
and
\begin{equation}
\bar{\Delta}(M)
=
\frac{1}{|\mathcal N|}\sum_{i\in\mathcal N}
\widehat{\Delta}_{i,\sigma_i(\hat t_i^M)}.
\label{eq:rz}
\end{equation}
These quantities are descriptive tools for comparing where a process evaluator places blame relative to the intervention-based estimate. They should not be interpreted as proof of attribution correctness unless the underlying step-level reference is itself validated.

For pre-execution forecasting, we evaluate both discrimination and calibration. Let $\widehat v_{0,i}$ denote the predicted success probability for instance $i$. We compute the Expected Calibration Error and Brier score as
\begin{equation}
\mathrm{ECE}
=
\sum_{b=1}^{B}\frac{|\mathcal{B}_b|}{n}
\left|\overline{y}_b-\overline{p}_b\right|,
\qquad
\mathrm{Brier}
=
\frac{1}{n}\sum_{i=1}^{n}\big(\widehat v_{0,i}-Y_i\big)^2.
\label{eq:calib}
\end{equation}
Because the benchmark has a high success base rate, we also report ranking-based or selective-risk quantities in the experiments to distinguish genuinely informative forecasts from nearly constant but calibrated predictors.

At the action level, next-action prediction is evaluated with top-$k$ accuracy, negative log-likelihood, calibration error, and effective candidate count. At the task level, uncertainty is evaluated through replay variance, residualized variance-structure probes, and its relationship to cheap predictive entropy from the action model. Together, these measurements allow us to answer three separate questions in the experiments: what predicts the next action, where uncertainty resides, and what process attribution methods actually measure. The full inventory of comparison baselines and nulls behind each measurement is given in Appendix~\ref{sec:baselines-app}.

\section{Experiments}
\label{sec:experiments}
\label{sec:experiment}

This section evaluates the proposed framework from the three-level perspective of the paper. Our goal is not only to report empirical results, but to determine what can be measured reliably at the action, task, and step levels of coding-agent execution. We therefore organize the experiments around five research questions, each tied to specific figures, tables, and comparison baselines.

We study the following questions. RQ1 asks which structural signal best predicts the next action of a coding agent. RQ2 asks what role the code dependency graph actually plays in execution. RQ3 examines whether commonly reported reuse patterns reflect genuine temporal behavior or permutation-invariant artifacts. RQ4 asks at which level execution uncertainty resides and whether step-level attribution is measurable at feasible replay cost. RQ5 asks what observational process attribution methods, especially full-trace judges, actually measure when no certified step-level causal reference is available.

\subsection{Experimental setup}
\label{sec:setup}
\label{sec:precond}

We study \NInstances{} file-localization instances derived from SWE-bench-Verified, covering \NRepos{} repositories. Each episode contains a natural-language issue, a repository snapshot, an agent trajectory, and a gold file set $F^\star$. The primary task outcome is recall over the gold file set, which we use because it preserves substantially more intervention-sensitive variation than binary success. Under this outcome, \PctMoved{} of instances can move under intervention, compared with only \PctMovedY{} under the binary outcome. Appendix~\ref{sec:byrepo} gives the per-repository composition, which is dominated by \texttt{django} and is why leave-one-repository-out is used wherever a predictor is fitted, and Appendix~\ref{sec:settings} gives the agent, model, and environment-pinning details.

The causal analysis relies on replayability and prefix observability, so we verify these preconditions empirically. Environment replayability holds at \ReplayRate{} over \ReplayN{} actions with \ReplayRepeats{} re-executions per action, and context compaction occurs in \NoCompaction{} of all rollouts. For attribution and uncertainty estimation, we replay \MReplay{} continuations from each of \NSeg{} prefixes across \NAttrPrime{} instances, yielding \ForkPrime{} replayed continuations over \HoursPrime{} hours. Confidence intervals are bootstrap intervals at the episode or instance level; null comparisons use within-episode permutation tests or closed-form expectations where available; and multiple comparisons are controlled by BH-FDR at $q=0.10$. Appendix~\ref{sec:baselines-app} lists every comparison predictor and null used below, and Appendix~\ref{sec:conventions-app} states the statistical conventions in full.

\subsection{RQ1: What predicts the next action?}
\label{sec:rq1}

\begin{table}[t]
\centering
\footnotesize
\setlength{\tabcolsep}{3pt}
\renewcommand{\arraystretch}{1.0}
\begin{tabularx}{\linewidth}{@{}>{\raggedright\arraybackslash}Xrrrrrrr@{}}
\toprule
\textbf{Object model} & \textbf{cov.} & \textbf{top-1} & \textbf{top-3} &
\textbf{top-10} & \textbf{NLL} & \textbf{ECE} & \textbf{$\exp(H)$} \\
\midrule
Routed by predicted tool & 0.619 & 0.150 & 0.260 & 0.484 & 2.76 & 0.0123 & 12.8 \\
Factored, shared object model & 0.645 & 0.109 & 0.241 & 0.432 & 3.41 & 0.0139 & 39.9 \\
\midrule
\textit{Paired difference} & --- & +0.041 & +0.018 & +0.052 & --- & --- & --- \\
\textit{95\% CI} & --- & [+.034, +.047] & [+.009, +.026] & [+.047, +.058] & --- & --- & --- \\
\bottomrule
\end{tabularx}
\caption{Predicting the \emph{complete} next action $(\mathrm{tool}, \mathrm{obj})$ over \NPts{} decision points from \NEpEval{} episodes. Both models share the same tool component and the same denominator: every decision point counts, and points whose realized action falls outside the candidate set are scored as misses rather than dropped. top-$k$ are closed-form expectations over random tie-breaking. NLL is over covered points only, so it measures sharpness rather than coverage. $\exp(H)$ is the median effective candidate count. Routing the \emph{construction} of the candidate set on the predicted next tool---reads take paths, searches take free-text queries, handler tools take no object---is what the dependence $I=\ToolObjMI{}$ nats ($p=\ToolObjMIP{}$) between tool and object type requires; it improves every top-$k$ and tightens the candidate set $3.1\times$.}
\label{tab:joint}
\end{table}

\begin{table}[t]
\centering
\small
\begin{tabular}{@{}lrrr@{}}
\toprule
\textbf{Stratum (observable before predicting)} & \textbf{$n$} &
\textbf{top-3} & \textbf{ECE} \\
\midrule
Current step is a search & 228 & 0.469 & 0.0125 \\
Current step is a read & 206 & 0.160 & 0.0247 \\
Current step lists a directory & 35 & 0.057 & 0.0278 \\
Predicted $\exp(H) \le 5$ & 154 & 0.351 & 0.0762 \\
Predicted $\exp(H) \in (5,15]$ & 198 & 0.409 & 0.0168 \\
Predicted $\exp(H) > 15$ & 157 & 0.159 & 0.0157 \\
Last output lists 1--3 paths & 263 & 0.357 & 0.0411 \\
Last output lists 4--10 paths & 151 & 0.298 & 0.0050 \\
Last output lists $\ge 11$ paths & 95 & 0.221 & 0.0125 \\
\bottomrule
\end{tabular}
\caption{When the next-action distribution can be trusted. Only strata
observable \emph{before} the prediction are listed, because a stratum defined
by the realized answer is a conditional accuracy and not a rule: conditioning on
coverage gives top-3 $=0.611$ at ECE $=0.0103$, which is not
actionable and which we therefore exclude. Two readings are practical. Prediction
is trustworthy immediately after a search and untrustworthy after a read, an
eight-fold spread. And low entropy is \emph{not} sufficient for trust: the
$\exp(H)\le 5$ stratum is the worst calibrated, because a small candidate set
signals coverage risk rather than confidence.}
\label{tab:trust}
\end{table}

\begin{table}[t]
\centering
\small
\setlength{\tabcolsep}{4pt}
\begin{tabular}{@{}lrrlrr@{}}
\toprule
\textbf{Next-object predictor} & \textbf{cov.} & \textbf{top-3} &
\textbf{95\% CI} & \textbf{top-10} & \textbf{$\exp(H)$} \\
\midrule
\multicolumn{6}{@{}l@{}}{\textit{Repository structure}} \\
\quad Repository frequency (marginal) & 0.349 & 0.072 & [0.047, 0.099] & 0.163 & 97.62 \\
\quad Code dependency graph, strengthened & 0.677 & 0.058 & [0.039, 0.078] & 0.172 & 1719.13 \\
\addlinespace[2pt]
\multicolumn{6}{@{}l@{}}{\textit{Visit history}} \\
\quad Visit-history frequency & 0.000 & 0.000 & [0.000, 0.000] & 0.000 & 4.84 \\
\addlinespace[2pt]
\multicolumn{6}{@{}l@{}}{\textit{Execution provenance}} \\
\quad Uniform over mentioned paths & \textbf{0.689} & 0.136 & [0.116, 0.155] & 0.354 & 25.00 \\
\quad Frequency-weighted history & \textbf{0.689} & 0.194 & [0.162, 0.227] & 0.431 & 22.17 \\
\quad Recency-weighted history & \textbf{0.689} & 0.308 & [0.267, 0.348] & \textbf{0.499} & 18.75 \\
\quad Last output only & 0.501 & 0.297 & [0.256, 0.338] & 0.431 & 4.00 \\
\quad Last output, position-ranked & 0.501 & \textbf{0.326} & [0.276, 0.375] & 0.455 & \textbf{3.46} \\
\addlinespace[2pt]
\multicolumn{6}{@{}l@{}}{\textit{Tool-routed candidate sets}} \\
\quad Routed on provenance & \textbf{0.689} & 0.156 & [0.126, 0.188] & 0.409 & 21.64 \\
\quad Routed on the code graph & 0.677 & 0.009 & [0.005, 0.014] & 0.109 & 283.28 \\
\bottomrule
\end{tabular}
\caption{The RQ1 predictor suite on the 341 \emph{first-visit} decision
points, where the realized object has never been visited before. This is the
discriminating regime: visit history covers none of these steps by construction,
so a predictor cannot score by restating the agent's own past, which is why the
visit-history row sits at zero coverage. All rows share one denominator---points
whose realized object falls outside the candidate set are scored as misses rather
than dropped---and top-$k$ are closed-form expectations over random tie-breaking.
The code-graph row is \emph{strengthened} relative to the obvious
implementation: it seeds a multi-source proximity ranking from the entire visited
set rather than from the last step alone, and it still loses to every provenance
variant listed, including the uninformative uniform one. The paired contrast
quoted in \S\ref{sec:rq1} uses the frequency-weighted row rather than the best
row, so it understates the gap. Higher is better for coverage, top-3
and top-10; lower is better for the effective candidate count $\exp(H)$. Best
value per column in bold.}
\label{tab:rq1baselines}
\end{table}

\begin{figure}[t]
\centering
\includegraphics[width=\linewidth]{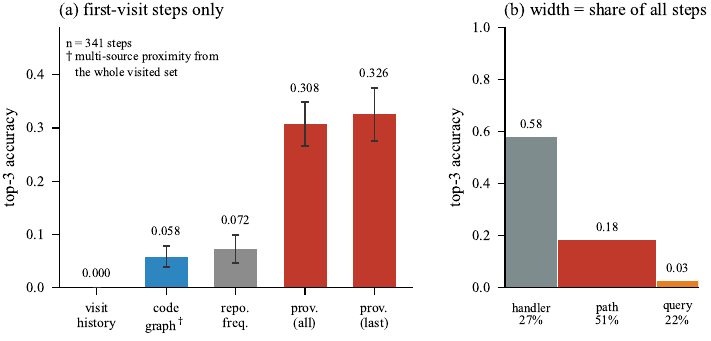}
\caption{\textbf{Execution provenance versus code structure on first-visit
steps}. The structure that predicts an agent's next object is its own execution
provenance, not the code dependency graph. (a) On the
\FirstVisitShare{} of steps whose object has never been visited---where visit
history is empty by construction---a predictor reading only the most recent tool
output reaches top-3 $=\ProvTopThree{}$, against $\GraphTopThree{}$ for a code
graph strengthened to seed a multi-source proximity ranking from the entire
visited set, and $\FreqTopThree{}$ for repository frequency. Bars are closed-form
expectations over random tie-breaking; whiskers are episode-level bootstrap 95\%
CIs. (b) Predictability is heterogeneous, and bar width is the share of all
\NPts{} steps: handler steps are nearly determined, path steps are moderately
predictable, and steps taking a novel free-text search query are not. A single
aggregate number would hide this.}
\label{fig:provenance}
\end{figure}

RQ1 asks which structural signal best predicts what the agent does next. The main competitors are execution provenance and repository structure. We evaluate this question with Figure~\ref{fig:provenance}, Table~\ref{tab:joint}, and Table~\ref{tab:trust}.

Figure~\ref{fig:provenance}(a) gives the central result. On the \FirstVisitShare{} of steps that move to a previously unvisited object, a provenance-only predictor using the most recent tool output attains top-3 $=\ProvTopThree{}$ with confidence interval $\ProvTopThreeCI{}$, compared with $\GraphTopThree{}$ for a strengthened dependency-graph predictor and $\FreqTopThree{}$ for a repository-frequency baseline. The paired improvement over the graph predictor is $\ProvVsGraphThree{}$ at top-3 with confidence interval $\ProvVsGraphThreeCI{}$, and $\ProvVsGraphTen{}$ at top-10 with confidence interval $\ProvVsGraphTenCI{}$. This comparison is especially informative because first-visit steps remove visit history by construction, so the gain cannot be explained by trivial revisitation effects. Instead, it shows that the agent's next move is guided primarily by what it has just seen in tool outputs.

The same figure also reveals why this effect holds. Among first-visit targets, \ProvCoverage{} of them had already been mentioned in earlier tool outputs, with confidence interval $\ProvCoverageCI{}$. In other words, execution provenance already covers most of the objects that later become next-step targets. This is a much stronger signal than graph proximity, which remains weak even after strengthening the graph baseline. The candidate efficiency numbers make the same point: provenance achieves its top-3 performance with only \ProvEff{} effective candidates, while broader graph-based search is less accurate despite a more diffuse candidate set.

Figure~\ref{fig:provenance}(b) shows that next-action predictability is strongly heterogeneous by object type. Handler steps, which account for \ShareHandler{} of all steps, reach top-3 $=\TopThreeHandler{}$; path steps, which account for \SharePath{}, reach $\TopThreePath{}$; and steps whose object is a novel free-text query, which account for \ShareQuery{}, reach only $\TopThreeQuery{}$. This heterogeneity matters because a single aggregate accuracy would hide the fact that query generation, not path selection, is the hard part of next-action prediction. That query number survives a deliberate attempt to improve it, and a parser defect we found and measured rather than assumed moves it \emph{down} rather than up; Appendix~\ref{sec:query-app} reports the audit, because the direction of the defect determines how the claim should be read.

Table~\ref{tab:joint} strengthens this interpretation by showing that the next action should be modeled jointly rather than with an independent factorization over tool and object. Tool identity and object type are statistically dependent, with mutual information $I=\ToolObjMI{}$ nats and permutation $p=\ToolObjMIP{}$. A routed predictor that constructs the candidate set conditional on the predicted tool improves top-1 from \FactoredTopOne{} to \RoutedTopOne{}, a gain of \RoutedGainOne{} with confidence interval \RoutedGainOneCI{}, and improves top-10 by \RoutedGainTen{} with confidence interval \RoutedGainTenCI{}. It also shrinks the effective candidate set from \FactoredEff{} to \RoutedEff{} while preserving calibration. This confirms that action prediction is structured not only by provenance but also by a nontrivial interaction between tool choice and object type.

Table~\ref{tab:trust} turns these findings into an actionable reliability analysis. Immediately after a search step, the next action is both accurate and calibrated, with top-3 $=\AfterSearchThree{}$ and ECE $=\AfterSearchECE{}$. After a read step, the same prediction problem becomes much harder, with top-3 only $\AfterReadThree{}$. Moreover, low predicted entropy is not by itself a reliable trust signal: the low-effective-candidate regime is in fact the worst calibrated, because a small candidate set often reflects coverage failure rather than genuine confidence.

Taken together, these results answer RQ1 clearly: the dominant predictor of a coding agent's next action is its own execution provenance, not the code dependency graph. Repository structure matters much less for local next-step prediction than what the agent has just observed.

\subsection{RQ2: What is the code dependency graph actually useful for?}
\label{sec:rq2}

\begin{figure}[t]
\centering
\includegraphics[width=\linewidth]{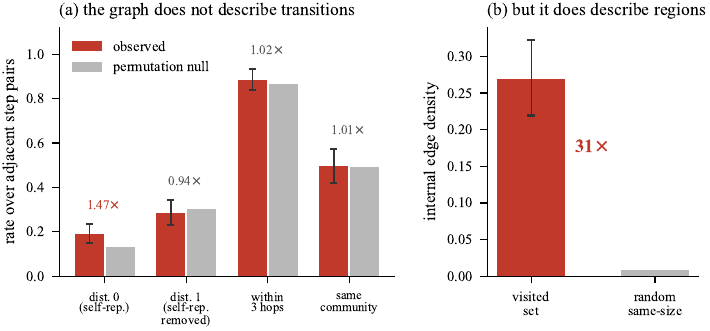}
\caption{\textbf{What the dependency graph does and does not describe}. The graph
describes where an agent works, not how it moves.
(a) Against a within-episode permutation null---which preserves the
multiset of visited files and destroys only their order---consecutive steps are
no closer in the graph than chance (one-hop lift $\OneHopLift{}$ with
self-repeats removed, three-hop $\ThreeHopLift{}$, same-community
$\CommunityLift{}$). The only quantity above chance is distance zero, i.e.\
immediate self-repetition, which is a temporal effect (\S\ref{sec:rq3}). (b) The
same graph separates the visited set from random same-size sets by
$\ClusterLift{}\times$ in internal edge density. The two panels are consistent:
inside a region of density $\ClusterObs{}$ an arbitrary pair is already likely to
be one hop apart, and the permutation null removes exactly that region effect.}
\label{fig:graph_null}
\end{figure}

RQ2 asks whether the code dependency graph is useless, or merely useful at a different level than local action prediction. We answer this with Figure~\ref{fig:graph_null} and the associated graph-ranking statistics.

Figure~\ref{fig:graph_null}(a) evaluates whether the graph explains \emph{step-to-step movement}. The null here is important: we compare the observed trajectory order against a within-episode permutation null that preserves the visited set but destroys only temporal order. Against this null, consecutive steps are not meaningfully closer in the graph than chance. The one-hop lift is $\OneHopLift{}$, the three-hop lift is $\ThreeHopLift{}$, and the same-community lift is $\CommunityLift{}$. The only quantity that clearly exceeds chance is immediate self-repetition, with lift $\SelfRepeatLift{}$, but this is a temporal reuse effect rather than evidence that the graph determines local transitions. This result directly complements RQ1: if provenance predicts next-step behavior well and graph proximity does not outperform a permutation null, then the graph should not be interpreted as the local transition kernel of agent navigation.

Figure~\ref{fig:graph_null}(b) shows a different role for the same graph. The visited set has internal edge density $\ClusterObs{}$, compared with a null of only $\ClusterNull{}$, corresponding to a lift of $\ClusterLift{}\times$. This means that the files touched by the agent form a graph-coherent region even though the order of movement inside that region is not graph-driven. The graph also ranks unseen gold files highly once leakage is controlled: the leakage-controlled rank quantile for unvisited gold files is \GoldRankUnseen{} with confidence interval \GoldRankUnseenCI{}, and remains \GoldRankPrefix{} when seeded from an early prefix. These quantities show that the graph is informative about which area of the repository matters, even if it is not informative about which file comes next.

The two findings are therefore complementary rather than contradictory. The graph describes the semantic region in which the agent is working, but not the temporal ordering of its steps within that region. RQ2 thus answers that the code dependency graph is useful as a scoping prior, not as a next-step transition model. This conclusion is confined to the \GraphResolveRate{} of agent objects that resolve to a graph node, a coverage limit we return to in \S\ref{sec:limitations}.

\subsection{RQ3: Which reuse patterns are real?}
\label{sec:rq3}

RQ3 asks which reuse phenomena reflect real temporal behavior and which are artifacts of the visited multiset. This distinction matters because reuse statistics are often interpreted as evidence of agent tendency without checking whether they are permutation-sensitive.

We begin with reuse rates. For both tools and objects, the raw probability of reusing a previously seen element is largely determined by the multiset of visited elements and therefore does not, by itself, reveal a temporal preference. This means that the observed tool reuse rate of \ToolReuseRate{} and object reuse rate of \ObjReuseRate{} are not informative as behavioral evidence in isolation. They are arithmetic consequences of vocabulary size and trajectory length.

What does carry behavioral content is temporal organization. Objects are repeated in immediately adjacent steps $\ObjSelfRepeat{}\times$ more often than expected under within-episode permutation, with permutation $p\sim\ObjSelfRepeatP{}$. Successive visits to the same object are also temporally clustered, with inter-visit gaps compressed to a ratio of $\GapRatio{}$ relative to chance. Among revisit events, the target's previous frequency exceeds the closed-form uniform expectation by $\FreqBias{}\times$ with confidence interval $\FreqBiasCI{}$, and its recency rank exceeds expectation by $\RecencyBias{}\times$ with confidence interval $\RecencyBiasCI{}$. These are genuine temporal effects and indicate that agents preferentially stay near recently or frequently used objects.

The same conclusion does not hold for tools. Tool self-repetition has lift only $\ToolSelfRepeat{}$ with $p=\ToolSelfRepeatP{}$, which is not significant. Instead, the tool channel exhibits structured transitions, with mutual-information lift $\ToolMILift{}\times$ chance and permutation $p\sim\ToolMIP{}$. This matches the action-prediction results in RQ1: tool usage is predictable, but not because agents blindly repeat the same tool.

RQ3 therefore distinguishes two notions that are often conflated. Reuse \emph{rates} are often uninformative, while reuse \emph{patterns}---especially immediate object repetition, temporal clustering, and recency/frequency bias---capture genuine execution behavior.

\subsection{RQ4: At which level does execution uncertainty live?}
\label{sec:rq4}
\label{sec:levels}

\begin{figure}[t]
\centering
\includegraphics[width=\linewidth]{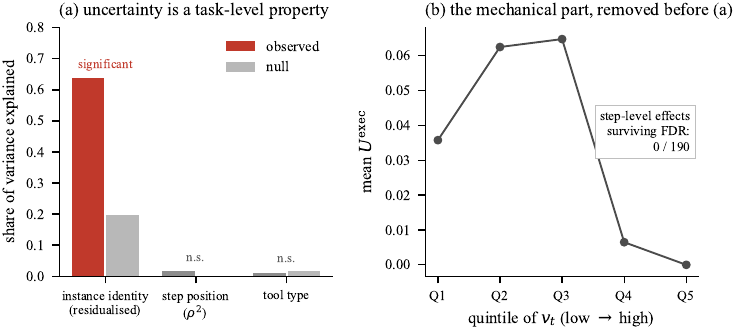}
\caption{\textbf{The level at which execution uncertainty resides}. Execution
uncertainty is a property of the task, not of the step.
(a) Probes for structure in the replay variance $U^{\mathrm{exec}}$,
all expressed as a share of variance explained. Instance identity explains
$\ICCResid{}$ against a permutation null of $\ICCResidNull{}$
($p=\ICCResidP{}$); step position and tool type explain nothing. (b) The
mechanical mean--variance relation that had to be removed before (a) is
credible: $U^{\mathrm{exec}}$ collapses to zero in the top quintiles of $v_t$,
because variance is constrained by the mean. Panel (a) is computed on residuals
after binning on $v_t$. Given the absence of step-level structure, the failure of
step-level effects to survive FDR correction is expected rather than
remediable by a denser scan.}
\label{fig:levels}
\end{figure}

RQ4 addresses the main empirical question behind the paper: whether the uncertainty relevant to execution outcome is localized at the step level or structured more globally at the task level. Figure~\ref{fig:levels} provides the core evidence, and the attribution statistics show the consequence for step-level measurement.

Figure~\ref{fig:levels}(a) compares several structural probes of replay variance. The key null result is that zero-cost predictive entropy from the action model does not proxy replay-based execution uncertainty. The correlation is $\rho=\HUexecRho{}$ at the step level with confidence interval $\HUexecCI{}$, and $\rho=\HUexecInstRho{}$ after aggregation to the instance level with confidence interval $\HUexecInstCI{}$. Neither indicates a useful relationship. This rules out a convenient but misleading interpretation in which cheap action uncertainty could stand in for actual execution variance. A control quantity does respond at $\rho=\RecallUexecRho{}$ with confidence interval $\RecallUexecCI{}$, so the null is not an artifact of the outcome measure.

At the same time, the replay variance is not structureless. Figure~\ref{fig:levels}(a) shows that instance identity explains a substantial share of residual variance after removing the mechanical mean--variance relationship, with ICC $=\ICCResid{}$ compared against a permutation null of $\ICCResidNull{}$ and $p=\ICCResidP{}$. By contrast, step position contributes little, with $\rho=\UexecPosRho{}$, and tool type shows no effect, with $p=\UexecToolP{}$. Figure~\ref{fig:levels}(b) explains why residualization is necessary in the first place: variance is mechanically constrained by the mean, and replay variance collapses near the ends of the bounded outcome range. Once this mechanical relation is removed, the remaining signal is clearly task-level rather than step-level. One apparent exception to this null, produced by thresholding a continuous response at $p<0.05$, does not replicate and decays monotonically under a threshold sweep; Appendix~\ref{sec:binary-app} reports it, because dichotomising a continuous quantity produces clean-looking strata at small $n$.

This task-level locus of uncertainty has a direct consequence for attribution. Of \NSeg{} intervention-estimated step effects, only \StepSigRaw{} reach uncorrected $p<0.05$, indistinguishable from the nominal chance rate, and \StepSigBH{} survive BH-FDR at $q=0.10$. The replay budget behind this result is already substantial---\ForkPrime{} replayed continuations over \HoursPrime{} hours---so the negative result is informative. It shows that formal identification does not automatically imply empirical measurability at the desired resolution. The limiting factor is not only engineering, but the location of uncertainty itself. We also report the coarseness of the instrument: at $m=\MReplay{}$ the median replay variance is zero and \UexecZeroShare{} of segments have variance exactly zero, while \DegenRate{} of episodes are degenerate in the sense that every estimated contribution is identically zero.

RQ4 therefore gives the core answer of the paper: execution uncertainty resides primarily at the task level rather than at the step level, and this is why step-level causal attribution remains underpowered at feasible replay budgets.

\subsection{RQ5: What do observational attribution methods actually measure?}
\label{sec:rq5}
\label{sec:judge}

\begin{figure}[t]
\centering
\includegraphics[width=\linewidth]{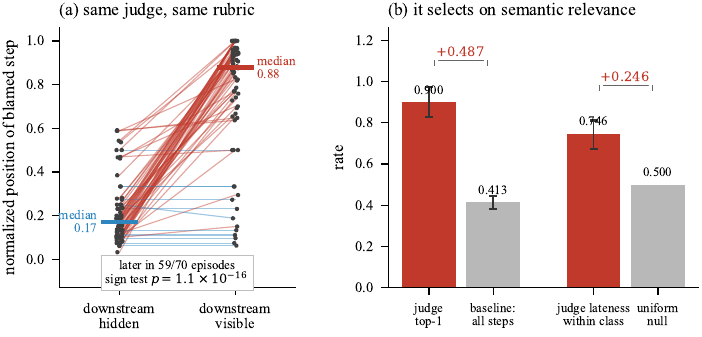}
\caption{\textbf{Collider isolation in full-trace judging}. Conditioning on
descendants of the scored step displaces the verdict later, and the manipulation
identifies this by construction rather than by inference.
(a) Each line is one episode. Judge model and rubric are fixed; the only
difference is whether steps downstream of the scored step are visible. The
verdict moves later in \CollLaterCount{} of \NAttr{} episodes
(\ColliderTied{} ties), by $\ColliderIso{}$ in mean normalized position. Because
the manipulation is of the judge's information set rather than of the world, this
requires no causal reference---which matters here, since none can be certified
(\S\ref{sec:rq4}). (b) What the full-trace judge selects on: steps that operate
on or reveal a gold file, at $\JudgeGold{}$ against the correct baseline of
$\GoldStepShare{}$, and biased late within that class against a closed-form
uniform null. That is semantic relevance---cheap and reproducible, but not
demonstrably causal contribution.}
\label{fig:collider_rq}
\end{figure}

\begin{table}[t]
\centering
\small
\begin{tabular}{@{}lrrrc@{}}
\toprule
\textbf{Zero-LLM structural rule} & \textbf{agree} &
\textbf{chance} & \textbf{diff} & \textbf{sig.} \\
\midrule
First step revealing a gold file & 0.043 & 0.053 & -0.010 & --- \\
First step operating on a gold file & 0.100 & 0.053 & +0.047 & --- \\
\textbf{Last} gold-provenance step & 0.243 & 0.053 & +0.190 & \checkmark \\
Last step of the trace (position only) & 0.171 & 0.053 & +0.118 & \checkmark \\
Step revealing the most gold files & 0.043 & 0.053 & -0.010 & --- \\
\bottomrule
\end{tabular}
\caption{Can a rule with no language model reproduce the full-trace judge's
choice? Agreement is exact same-step agreement over \NAttr{} episodes; chance is
the closed-form expectation $1/T_i$ for a rule selecting uniformly among steps.
The judge's \emph{class} of choice is almost fully explained---it selects
gold-provenance steps at \JudgeGold{} against a \GoldStepShare{} baseline, and
is biased late within that class ($\WithinLate{}$ against a uniform null of
$0.5$). Its exact choice is not: with $\NGoldSteps{}$ gold-provenance steps per
episode the within-class chance rate is $\WithinChance{}$, so the best rule is
only $1.7\times$ better than guessing inside the class. We report this residual
rather than presenting the judge as a lookup table.}
\label{tab:judge}
\end{table}

\begin{table}[t]
\centering
\footnotesize
\setlength{\tabcolsep}{3pt}
\renewcommand{\arraystretch}{1.0}
\begin{threeparttable}
\begin{tabularx}{\linewidth}{@{}>{\raggedright\arraybackslash}Xcccccc@{}}
\toprule
& \multicolumn{4}{c}{\textbf{Collider displacement}} &
\multicolumn{2}{c}{\textbf{Gold-prov. enrichment}} \\
\cmidrule(lr){2-5} \cmidrule(lr){6-7}
\textbf{Judge} & \textbf{mean} & \textbf{95\% CI} & \textbf{later} &
\textbf{sign $p$} & \textbf{rate} & \textbf{diff CI} \\
\midrule
gpt-5.5 (original)      & +0.533 & [+.399, +.656] & 19/23 & $4{\times}10^{-6}$ & 0.826$^{*}$ & [+.241, +.500] \\
qwen3.6-plus (general)  & +0.434 & [+.299, +.557] & 18/23 & $8{\times}10^{-6}$ & 0.565        & [-.068, +.297] \\
qwen3-coder-plus (code) & +0.394 & [+.251, +.551] & 15/23 & $6{\times}10^{-5}$ & 0.826$^{*}$ & [+.222, +.514] \\
\bottomrule
\end{tabularx}
\begin{tablenotes}[flushleft]
\footnotesize
\item Between-judge heterogeneity (permuting judge labels within episode):
displacement $p=0.181$, enrichment $p=0.044$.
\end{tablenotes}
\end{threeparttable}
\caption{The same manipulation across \NJudges{} judges on their \NCommonEp{}
common episodes. Displacement---the effect of making steps downstream of the
scored step visible---appears in every judge and shows no detectable
between-judge heterogeneity; the spread of $\DispSpread{}$ across judges is small
against per-judge intervals of roughly $\pm0.13$, which is what supports the
claim, since a $p$ of $0.181$ on \NCommonEp{} episodes means only that no
difference was detected. Enrichment behaves oppositely: it reaches significance in
\JudgesEnriched{} of \NJudges{} judges ($^{*}$) and its heterogeneity \emph{is}
significant. The baseline for the enrichment column is $\EnrichBaseCommon{}$, the
share of steps that touch or reveal a gold file. A critique of the mechanism
therefore transfers across scorers; a critique of what they score does not.}
\label{tab:crossmodel}
\end{table}

RQ5 asks what can still be said about observational attribution methods when no certified step-level causal reference is available. We answer this using Figure~\ref{fig:collider_rq}, Table~\ref{tab:judge}, and Table~\ref{tab:crossmodel}.

Figure~\ref{fig:collider_rq}(a) gives the central experiment. We hold the judge model and rubric fixed and vary only whether downstream steps are visible. If full-trace judges were implicitly approximating a prefix-conditioned step-local quantity, this manipulation should not systematically change the location of blame. Instead, the verdict moves $\ColliderIso{}$ later in normalized position, with confidence interval $\ColliderIsoCI{}$, and it moves later in \CollLaterCount{} of \NAttr{} episodes, with sign test $p=\ColliderSignP{}$. Because the world being judged is unchanged, this displacement isolates a bias in the evaluation mechanism itself rather than disagreement with a noisy causal reference. In causal terms, full-trace judging conditions on descendants of the scored step, and the later shift is exactly the empirical signature one would expect from such post-treatment adjustment.

Figure~\ref{fig:collider_rq}(b) then asks what the full-trace judge is selecting when it sees the whole trajectory. The answer is not arbitrary. The judge places \JudgeGold{} of its top-1 verdicts on steps that either operate on or reveal a gold file, compared with a baseline frequency of only \GoldStepShare{} among all steps. The enrichment over baseline is therefore \JudgeGoldDiff{} with confidence interval \JudgeGoldDiffCI{}. This means that the judge does capture semantically relevant evidence. However, within that semantically relevant class, it is biased late: the within-class lateness is \WithinLate{} against a uniform null of $0.5$, with difference \WithinLateDiff{} and confidence interval \WithinLateDiffCI{}. Thus, full-trace judging is neither random nor causally faithful. It preferentially selects semantically relevant steps, but it does so using information that is unavailable to a valid prefix-conditioned attribution rule.

Table~\ref{tab:judge} sharpens this point by asking whether a simple non-LLM structural rule can reproduce the judge's top-1 selection. The best such rule---choosing the last gold-provenance step---achieves agreement \BestRuleAgree{}, well above the closed-form chance level of $0.053$, but still far from perfect. This is a useful result for interpretation. It shows that the judge is not behaving like a trivial lookup rule, yet its behavior is largely captured by a preference for late, gold-provenance-related evidence. In other words, what the judge extracts is semantically meaningful, but not demonstrably causal.

We next compare this observational signal to the intervention-based estimate. On the estimation points available to SCAE, the intervention-based top-1 lands on gold-provenance steps at rate \CausalGold{}, against its own baseline of \GoldProvEst{}. Restricting to the subset of steps that are directly comparable between the judge and the intervention estimate gives \FairJudge{} versus \FairCausal{}, with difference \FairDiff{} and confidence interval \FairDiffCI{}; only \JudgeInEst{} of judge verdicts are step-comparable at all, which is why the two baselines must be computed separately. Because RQ4 showed that the intervention-based step reference does not survive FDR correction, we do not interpret this gap as proof that one selector is correct and the other is wrong. Instead, we interpret it more conservatively: the two procedures are measuring different quantities, and the judge's quantity is better understood as downstream-informed semantic relevance.

Table~\ref{tab:crossmodel} shows that the mechanism-level result generalizes across judge models. Repeating the same downstream-visibility manipulation with \NJudges{} judges over their \NCommonEp{} common episodes yields positive displacement in all cases: $\DispOrig{}$ for the original judge, $\DispQwen{}$ for the second general model, and $\DispCoder{}$ for the code-specialized model. Each judge's confidence interval excludes zero, and each sign test is below $10^{-4}$. Moreover, the heterogeneity test for displacement is not significant ($p=\HetDispP{}$), and the spread of only $\DispSpread{}$ across judges is small relative to the per-judge intervals. This supports the claim that collider displacement is a stable property of the judging mechanism rather than an idiosyncrasy of a single scorer.

At the same time, Table~\ref{tab:crossmodel} also shows that the \emph{content} of judging is more model-dependent. Only \JudgesEnriched{} of \NJudges{} judges exhibit significant enrichment on gold-provenance steps above the baseline of \EnrichBaseCommon{}, and the heterogeneity test for this enrichment is significant ($p=\HetEnrichP{}$). This contrast is important for the paper's broader argument. The mechanism-level critique transfers across scorers: full-trace judging is systematically biased by downstream visibility. The content-level interpretation does not transfer as cleanly, because different judge models rely on somewhat different semantic cues.

Finally, we revisit a pre-registered criterion that turns out to fail as a criterion. The judge/causal top-1 disagreement rate is \DisagreeRate{}, which might at first glance appear to support a strong attribution critique. However, once the intervention-based step reference is known not to survive FDR correction, this disagreement rate is no longer diagnostic. Under the null, a noisy selector over a small set of estimation points and a judge choosing from roughly twenty trajectory steps will disagree almost all the time. We therefore retire this criterion rather than treating it as positive evidence; Appendix~\ref{sec:metrics-app} works out the null and Appendix~\ref{sec:prereg} records the withdrawal. The visibility-manipulation contrast above is the correct replacement, because it identifies a property of the judge without relying on an uncertified causal reference.

Taken together, RQ5 shows that observational attribution methods in this setting do not provide validated step-level causal explanations. What they measure is better described as semantically enriched, downstream-informed relevance, and the controlled visibility experiment identifies the source of this mismatch directly.

\subsection{Summary of experimental findings}
\label{sec:exp_summary}

Across all five research questions, the results support the same three-level account of process evaluation. At the action level, agent behavior is strongly structured and often predictable, but mainly by execution provenance rather than repository transitions. At the task level, execution uncertainty is real and structured, but primarily at the instance level rather than the step level. At the step level, formal identifiability does not translate into practical measurability at feasible replay cost, and observational attribution methods instead rely heavily on downstream semantic evidence.

These findings explain why process evaluation can appear simultaneously useful and difficult to validate. Some process-level signals are useful because they predict what the agent will do next. Others are useful because they characterize whether the task remains uncertain. But those signals should not automatically be interpreted as step-level causal attribution. The experiments therefore support the central claim of the paper: action, task, and step are different levels, and process evaluators should be interpreted according to the level they actually measure. Appendix~\ref{sec:discussion} states, for each of these claims, the observation that would undercut it.

\section{Limitations}
\label{sec:limitations}

This paper makes a measurement claim about process evaluation in coding agents, and its conclusions should be read within the scope of that claim. The first limitation is task scope. Our analysis is conducted on file localization rather than the full space of coding-agent behaviors, because this setting provides the replayability, prefix reconstruction, and verifiable outcomes required by the method. File localization is therefore not a proxy chosen for convenience, but the smallest realistic environment in which the identification assumptions can be checked empirically. At the same time, this choice limits external validity: process structure in code editing, patch synthesis, or open-ended tool use may differ substantially from the structure observed here.

A second limitation is that the intervention-based step reference is formally identified but empirically weak. Of \NSeg{} intervention-estimated effects, \StepSigBH{} survive BH-FDR correction at $q=0.10$, and only \StepSigRaw{} reach uncorrected $p<0.05$, which is consistent with the nominal chance rate. This means that we cannot certify which step mattered in any individual episode. The strongest attribution result in the paper is therefore not a correctness claim about a judge's selected step, but a mechanism claim about how the judge's verdict changes when downstream information becomes visible. More broadly, our conclusions about step-level attribution should be read as limits on measurement at feasible replay cost rather than as proof that step-local causal structure never exists.

A third limitation concerns the replay budget and the granularity of estimation. We estimate values at selected prefixes rather than at every step in the trajectory, and each estimation point is backed by a finite number of replays. As a result, the reported contributions are segment-level approximations rather than dense per-step measurements. Increasing replay density would improve resolution, but only up to the point allowed by the structure of the underlying uncertainty. Since our results place that uncertainty mainly at the task level rather than the step level, simply scaling replay count is unlikely to remove the core limitation by itself.

A fourth limitation is that several assumptions are only partially testable. Replayability and compaction can be checked empirically in the current harness, and we do so. By contrast, exogeneity of decoding noise is a structural assumption about the serving policy and cannot be verified directly from traces. Similarly, the intervention branch relies on a transferability assumption about the plausibility of substituted actions under a perturbed policy. These assumptions are explicit in the method, but they remain assumptions, and future work should test how sensitive the conclusions are to departures from them.

A fifth limitation concerns judge evaluation. We replicate the collider-displacement finding across \NJudges{} judge models and find no detectable heterogeneity in displacement, but the cross-model sample remains modest and the heterogeneity test is necessarily low-powered. The content-level characterization of what judges score is even less stable across models than the displacement mechanism itself. Our strongest claim therefore concerns the information-set effect in full-trace judging, not a complete universal account of all LLM-based process evaluators.

A sixth limitation is that the uncertainty decomposition is only partially validated on real traces. The policy-driven component is directly estimated from repeated executions, but the full three-way decomposition into task, policy, and model uncertainty is validated only on a synthetic known-generating process. In the main benchmark setting, we use this decomposition as an organizing framework for uncertainty analysis rather than as a fully solved inference problem. The empirical claim supported by the experiments is narrower and more robust: the uncertainty that matters for execution outcome is primarily structured at the task level.

Finally, the graph-based analyses are limited by structural coverage. Only \GraphResolveRate{} of agent objects resolve cleanly to nodes in the dependency graph, and free-text search queries remain only weakly represented by graph structure. This does not affect the main provenance-based findings, but it does limit how strongly graph-based conclusions can be generalized beyond the resolvable subset.

These limitations do not undermine the central measurement result of the paper, but they define its proper scope. Our claim is not that process evaluation is useless, nor that step-level causal attribution is impossible in every setting. Our claim is that, in a replayable coding-agent benchmark where the causal target can be defined precisely, action prediction, task uncertainty, and step attribution behave differently enough that they should not be treated as a single evaluative object.

\section{Conclusion}
\label{sec:conclusion}

Process evaluation is increasingly used to analyze coding agents beyond final task success, but existing practice often treats all process-level signals as if they supported the same kind of interpretation. This paper argues that such a view is too coarse. Action prediction, task uncertainty, and step attribution are different levels of analysis, and conclusions drawn at one level do not automatically transfer to the others.

To study this problem, we introduced a measurement framework centered on SCAE, a replay-based estimator derived from a structural causal model of agent execution. This framework allowed us to define a prefix-conditioned step-level causal quantity, estimate it under replay and intervention, and compare it against observational process evaluators under controlled changes to their information sets. Using this instrument, we asked three connected questions: what predicts the next action, where execution uncertainty resides, and what full-trace attribution methods actually measure.

The experiments yield three main findings. First, next actions are predicted primarily by execution provenance rather than by code-graph transitions, which shows that agents move mainly according to what they have just observed rather than according to repository structure alone. Second, execution uncertainty has meaningful instance-level structure but little step-level structure, which implies that the uncertainty most relevant to outcome is a property of the task rather than of individual steps. Third, full-trace judges shift their selected blame later when downstream steps become visible, isolating collider bias in the evaluation mechanism and showing that observational attribution is better understood as semantically enriched, downstream-informed relevance than as certified causal contribution.

Taken together, these findings explain a tension that has become common in process evaluation. Process-level signals can be useful and still fail to support the interpretation often assigned to them. Action-level predictors can be strong, task-level uncertainty can be structured, and full-trace judges can select semantically meaningful steps, while step-level causal localization remains infeasible at practical replay budgets. This is not a contradiction once the levels are separated; it is the main reason the paper insists on separating them.

The broader implication is methodological. Process evaluation should not be treated as a single benchmark target whose meaning is fixed in advance. Instead, evaluation methods should be validated against the level they actually measure. We hope this work encourages future research to build process evaluators whose targets are explicit, whose assumptions are testable, and whose outputs are interpreted according to the kind of evidence they genuinely provide.

\clearpage
\bibliographystyle{iclr2027_conference}
\bibliography{iclr2027_conference}

\clearpage
\appendix

\section{Notation and Formal Material}
\label{sec:formal-app}

This appendix supplies the formal material behind \S\ref{sec:method}: the notation
used throughout (Table~\ref{tab:notation}), the consequence of each conditioning
set for current practice, where on the causal ladder our estimates sit, the three
ways identification can fail, and the proofs of Proposition~\ref{prop:seqrand} and
Theorem~\ref{thm:additive}.

\subsection{Notation}
\label{sec:notation}

\begin{table}[h]
\centering
\small
\begin{tabular}{@{}ll@{}}
\toprule
\textbf{Symbol} & \textbf{Definition} \\
\midrule
$q$ & natural-language issue text of an episode \\
$G = (V,E)$ & repository graph over files, modules and symbols \\
$F^\star$ & gold target set: files modified by the reference patch \\
$\hat F$ & file set predicted by the agent \\
$Y$ & task-level outcome $\mathbb{1}[F^\star \subseteq \hat F]$ \\
$Y_f$, $Y_{\mathrm{rec}}$ & per-file indicator; recall $|F^\star \cap \hat F| / |F^\star|$ \\
$A_t$, $O_t$ & action taken at step $t$; tool output it returns \\
$S_t$ & agent state at step $t$; $S_1 = \sigma(q,\mathrm{sys})$ \\
$\tau_{1:t}$ & trace prefix $(A_1,O_1,\dots,A_t,O_t)$ \\
$T$ & number of steps in an episode; $t/T$ is normalized position \\
$Z$ & latent task difficulty (exogenous) \\
$G^{(t)}_\pi$ & decoding noise at step $t$, $\mathrm{Gumbel}(0,1)$ (exogenous) \\
$U^{(t)}_{\mathrm{env}}$ & environment noise at step $t$ (exogenous, pinned) \\
$U^{(\ge t)}$ & all exogenous noise from step $t$ onward \\
$\pi$, $\tilde\pi$ & deployed policy; raised-temperature perturbed policy \\
$\widehat H(\widehat\pi_t)$ & estimated action entropy at step $t$, Eq.~\eqref{eq:entropy} \\
$\tau_H$ & entropy threshold selecting the branch of Eq.~\eqref{eq:hybrid} \\
$m$ & replayed continuations per estimation point \\
$v_t$ & anytime value $\mathbb{E}[Y \mid \tau_{1:t}, q, G]$ \\
$v_0$ & pre-execution prediction $\mathbb{E}[Y \mid q, G]$ \\
$\Delta_t$ & step contribution $v_t - v_{t-1}$ \\
$U^{\mathrm{exec}}_t$ & replay variance of the outcome at estimation point $t$ \\
$\theta_t(a,a')$ & step-level causal contrast of Eq.~\eqref{eq:estimand} \\
$\mathrm{PN}_t$ & probability of necessity at step $t$ \\
$\mathrm{Infl}(t)$ & counterfactual influence at step $t$ \\
$U_{\mathrm{task}}$, $U_{\mathrm{policy}}$, $U_{\mathrm{model}}$ & the three components of Eq.~\eqref{eq:decomp3} \\
$\Sigma$ & Laplace posterior covariance of the $v_0$ model \\
$\mathcal{C}(q)$ & conformal prediction set for the target files \\
\bottomrule
\end{tabular}
\caption{Notation used throughout the paper. Symbols are introduced in
\S\ref{sec:method} and reused without redefinition in \S\ref{sec:experiments}.}
\label{tab:notation}
\end{table}

\subsection{Consequences of the conditioning set for current practice}
\label{sec:consequences}

Proposition~\ref{prop:seqrand} licenses exactly one conditioning set: the realized
prefix. Table~\ref{tab:consequences} states, for each departure from it that
appears in current practice, the resulting bias and the signature it should leave
in data. The mechanism in the first row is standard --- conditioning on a
descendant of the treatment opens a non-causal path and biases the estimate
\citep{pearl2009causality} --- but it is not usually applied to LLM-as-a-judge
attribution \citep{zheng2023judging}, which presents the entire trajectory and the
realized outcome and then asks which step was at fault. That row is also the only
one yielding a falsifiable prediction without requiring that our own estimator be
correct, which is why \S\ref{sec:rq5} tests it and why we foreground it.

\begin{table}[h]
\centering
\small
\begin{tabular}{@{}p{0.24\textwidth}p{0.22\textwidth}p{0.44\textwidth}@{}}
\toprule
\textbf{Practice} & \textbf{Conditions on} & \textbf{Consequence} \\
\midrule
Full-trace LLM judge &
prefix $+$ \emph{successors} $+$ outcome &
Successors are descendants of $A_t$, and conditioning on a descendant opens a
biasing path. Predicted signature: blame displaced toward \emph{later} steps. \\
Process reward models &
nominally the prefix &
Labels back-filled from completed trajectories reintroduce future information. \\
Ablation with restart &
nothing &
Resamples all noise, so it estimates a marginal interventional contrast and
discards the realized history. \\
Data-flow ``dead step'' analysis &
syntactic dependencies &
Conflates data-flow reachability with causal effect on the outcome
(\S\ref{sec:dataflow}). \\
\bottomrule
\end{tabular}
\caption{Each row misestimates Eq.~\eqref{eq:estimand} for a distinct reason, and
none of the four reasons is unmeasured confounding. Only the first row yields a
signature testable without a causal reference, which is why it is the attribution
result we foreground in \S\ref{sec:rq5}.}
\label{tab:consequences}
\end{table}

\subsection{The causal ladder, instantiated for localization}
\label{sec:ladder-app}

\begin{table}[h]
\centering
\small
\begin{tabular}{@{}llll@{}}
\toprule
\textbf{Level} & \textbf{Expression} & \textbf{Meaning for localization} & \textbf{Typical practice} \\
\midrule
I & $P(\hat F \mid q)$ & association between query and result & retrieval recall@$k$ \\
II & $P(\hat F \mid do(a'))$ & alter a step and \emph{restart} & tool-ablation studies \\
II.5 & $P(\hat F \mid \tau_{1:t-1}, do(a'))$ & fix realized history, alter one step & \textbf{SCAE (continuation)} \\
III & $P(\hat F_{a'} \mid \tau_{\mathrm{obs}})$ & fix realized noise, alter one step & \textbf{SCAE (intervened step)} \\
\bottomrule
\end{tabular}
\caption{Levels II and II.5 differ in whether the episode's own history is
retained. Most agent ablation studies operate at Level II while being described in
counterfactual language, which is the confusion the distinction is meant to
prevent. \S\ref{sec:ladder-position} states where the boundary falls for SCAE.}
\label{tab:ladder}
\end{table}

\subsection{Position on the causal ladder}
\label{sec:ladder-position}

With API-served models we obtain per-token log-probabilities only in a reference
arm; in the main arm we estimate $\widehat{\pi}_t$ over action equivalence classes
by resampling. The intervened step therefore admits Gumbel--max abduction and is
genuinely Level~III, whereas the continuation marginalizes residual policy noise
conditional on the prefix and is Level~II.5 in the terms of
Table~\ref{tab:ladder}. Environment noise is Level~III throughout, being pinned by
construction. We therefore claim action-level counterfactuals with a
prefix-conditioned continuation, and \emph{not} exact token-level
counterfactuals. Quantifying the gap requires the reference arm, whose
construction we describe but whose results are not included here.

\subsection{Where identification breaks}
\label{sec:breaks}

Three failure modes deserve naming, and none of them is confounding.

\paragraph{Positivity.}
This is the binding constraint in practice. Necessity-type quantities are the ones
it binds hardest on: the probability of necessity at step $t$,
\[
\mathrm{PN}_t = P\big(Y_{A_t=a^\star_t}{=}1 \;\big|\; \tau_{1:t-1}, A_t{=}a_t, Y{=}0\big),
\]
asks whether the outcome would have been correct had the agent taken the better
action $a^\star_t$, given that it took $a_t$ and failed. It is defined only when
$\pi(a^\star_t\mid\tau_{1:t-1}) > 0$. When the policy is effectively deterministic
at step $t$ this fails, and no amount of observational data recovers the quantity,
because the required comparison arm is never realized. This is why
\S\ref{sec:entropy-app} measures $\pi(\cdot\mid\tau_{1:t-1})$ rather than assuming
it, and why Eq.~\eqref{eq:hybrid} switches to a soft intervention exactly where the
measurement says overlap is absent.

\paragraph{Context compaction.}
Let $\varphi$ denote the harness's compaction map, so that the state actually
conditioned on is $\varphi(h_t)$ rather than $h_t$. When $\varphi$ discards
content, $h_t \mapsto \varphi(h_t)$ is not invertible and there exists
$D_t \notin \tau_{1:t-1}$ that influenced $A_t$. Assumption~\ref{as:a1} then
fails, Step~1 of the proof in \S\ref{sec:proofs} does not go through, and $D_t$
acts as an unmeasured confounder between $A_t$ and $Y$. The estimand is
unidentified, not merely harder to estimate: no reweighting of the observed trace
recovers it, because the confounder is absent from the log by construction. What
makes this tractable in principle is that compaction is triggered by the harness
rather than by nature, so it can be \emph{randomized} --- assigning compaction at
a fixed prefix converts a sensitivity-analysis problem into a controlled
experiment. We reserved the hook for this but did not run it, since no episode in
our sweep compacted spontaneously.

\paragraph{Counterfactual influence decay.}
Even with Gumbel--max abduction at the intervened step, the coupling between the
factual and counterfactual continuations weakens as the two paths diverge, so a
nominal Level~III quantity degrades toward a Level~II one over long horizons
\citep{kazemi2024counterfactual}. To keep this measurable rather than rhetorical we
define
\[
\mathrm{Infl}(t) = \mathrm{TV}\Big(P\big(\hat F\mid \tau_{\mathrm{obs}}, do(a')\big),\;
P\big(\hat F\mid do(a')\big)\Big),
\]
the total-variation distance between the counterfactual predicted-set distribution
that retains the observed trace and the one that discards it. $\mathrm{Infl}(t)$
near zero means the realized history has stopped constraining the continuation, so
the estimate should be reported as interventional rather than counterfactual.

\subsection{Proofs}
\label{sec:proofs}

Throughout, we read Assumption~\ref{as:a2} in its joint form: the decoding noises
$\{G^{(t)}_\pi\}_{t \ge 1}$ are mutually independent and independent of
$(Z, U_{\mathrm{env}})$. This is the standard exogeneity assumption for a sampler
whose temperature is a decoder parameter, and it is what ``exogenous'' asserts; we
make it explicit because the proof uses independence across steps, not only at
step $t$.

\subsubsection{Proof of Proposition~\ref{prop:seqrand}}

\paragraph{Step 1: conditional on the prefix, the action is a function of decoding
noise alone.}
Fix a prefix value $\tau_{1:t-1} = \tau$ with $P(\tau_{1:t-1} = \tau) > 0$. By
Assumption~\ref{as:a1} there is a deterministic map $f$ with $S_t = f(q, \tau)$,
so on the conditioning event $S_t$ takes a single fixed value. By
Eq.~\eqref{eq:action},
\[
A_t = \arg\max_k \big(\log \pi_k(f(q,\tau)) + G^{(t)}_{\pi,k}\big)
    =: g_\tau\big(G^{(t)}_\pi\big),
\]
a measurable function of $G^{(t)}_\pi$ only.

\paragraph{Step 2: that noise is independent of the prefix and of the continuation
noise.}
The prefix $\tau_{1:t-1}$ is, by the structural equations, a measurable function of
$\big(q, Z, \{G^{(s)}_\pi\}_{s<t}, \{U^{(s)}_{\mathrm{env}}\}_{s<t}\big)$. Write
$U^{(\ge t)} = \big(\{G^{(s)}_\pi\}_{s>t}, \{U^{(s)}_{\mathrm{env}}\}_{s\ge t}\big)$.
By Assumption~\ref{as:a2} in its joint form, $G^{(t)}_\pi$ is independent of the
collection $\big(Z, \{G^{(s)}_\pi\}_{s \ne t}, U_{\mathrm{env}}\big)$, hence
\begin{equation}
G^{(t)}_\pi \;\perp\; \big(Z,\; U^{(\ge t)},\; \tau_{1:t-1}\big).
\label{eq:noise-indep}
\end{equation}
Combining Eq.~\eqref{eq:noise-indep} with Step~1, and noting that conditioning on
$\tau_{1:t-1}$ leaves the law of $G^{(t)}_\pi$ unchanged,
\[
A_t \;\perp\; \big(Z,\; U^{(\ge t)}\big) \;\big|\; \tau_{1:t-1},
\]
which is the first claim.

\paragraph{Step 3: the potential outcome is a function of the continuation noise.}
Fix an action $a$. Recursively applying
$O_s = \mathcal{E}(A_s, G; U^{(s)}_{\mathrm{env}})$,
$S_{s+1} = \mathcal{T}(S_s, A_s, O_s)$ and $\hat F = \phi(S_T)$ from step $t$
onward, with $A_t$ set to $a$, the potential outcome under the intervention is a
deterministic function
\begin{equation}
Y_{A_t = a} \;=\; h\big(a,\; \tau_{1:t-1},\; U^{(\ge t)}\big),
\label{eq:po-form}
\end{equation}
where determinism of each $O_s$ given $(A_s, G, U^{(s)}_{\mathrm{env}})$ is exactly
Assumption~\ref{as:a3}. By Step~2 and Eq.~\eqref{eq:po-form},
$A_t \perp Y_{A_t=a} \mid \tau_{1:t-1}$ for every $a$, which is conditional
ignorability at step $t$.

\paragraph{Step 4: identification.}
Consistency holds by construction in a structural causal model: on the event
$\{A_t = a\}$ we have $Y = Y_{A_t=a}$. Therefore, for any $a$ with
$\pi(a \mid \tau_{1:t-1}) > 0$, so that the conditional expectation is well
defined,
\[
\begin{aligned}
\mathbb{E}[Y \mid \tau_{1:t-1}, do(A_t{=}a)]
&\overset{\text{(i)}}{=}
\mathbb{E}[Y_{A_t=a} \mid \tau_{1:t-1}] \\
&\overset{\text{(ii)}}{=}
\mathbb{E}[Y_{A_t=a} \mid \tau_{1:t-1}, A_t{=}a] \\
&\overset{\text{(iii)}}{=}
\mathbb{E}[Y \mid \tau_{1:t-1}, A_t{=}a].
\end{aligned}
\]
where (i) is the definition of the interventional expectation in $\mathcal{M}$,
(ii) is Step~3, and (iii) is consistency. This is Eq.~\eqref{eq:ident}, and it is
the one-step $g$-formula \citep{robins1986new} with a \emph{known} assignment
mechanism. \qed

\paragraph{Relation to dynamic treatment regimes.}
The sequential structure here is formally that of a dynamic treatment regime
\citep{robins1986new,murphy2003optimal}, but the difficulty is inverted. In
epidemiological applications the central obstacle is time-varying confounding: the
analyst must argue that everything the treatment decision depended on was
recorded. Here that argument is free, because the policy is a function of a context
we log in full and its decoding noise is a decoder property. What is \emph{not}
free is positivity, which those applications typically obtain from natural
variation and which we must measure and then work around
(\S\ref{sec:breaks}, \S\ref{sec:entropy-app}).

\paragraph{Remark on what the proof does and does not deliver.}
The positivity requirement in Step~4 is not a technicality but the binding
constraint in practice, which is what forces the hybrid estimator of
Eq.~\eqref{eq:hybrid}. Note also that Step~2 is where the argument fails for the
conditioning sets of Table~\ref{tab:consequences}: conditioning additionally on
$A_{t'}$ for $t' > t$ or on $Y$ conditions on functions of $U^{(\ge t)}$, which
breaks Eq.~\eqref{eq:noise-indep} and hence Step~3. Similarly, if the harness
compacts context then $S_t$ is not a function of $(q, \tau_{1:t-1})$,
Assumption~\ref{as:a1} fails, and Step~1 does not go through.

\subsubsection{Proof of Theorem~\ref{thm:additive}}

\paragraph{Telescoping.}
By definition $\Delta_t = v_t - v_{t-1}$, so
$\sum_{t=1}^T \Delta_t = v_T - v_0$ as a finite telescoping sum. The outcome
$Y = \mathbb{1}[F^\star \subseteq \phi(S_T)]$ is a deterministic function of
$S_T$, which by Assumption~\ref{as:a1} is a function of $(q, \tau_{1:T})$;
therefore $v_T = \mathbb{E}[Y \mid \tau_{1:T}, q, G] = Y$ almost surely, and
$v_0 = \mathbb{E}[Y \mid q, G]$ by definition. This gives
Eq.~\eqref{eq:telescope}.

\paragraph{The action/observation decomposition.}
Insert the intermediate conditioning set $(\tau_{1:t-1}, A_t{=}a_t)$, which lies
between $\tau_{1:t-1}$ and $\tau_{1:t} = (\tau_{1:t-1}, a_t, o_t)$:
\begin{equation}
\Delta_t = \underbrace{\Big(\mathbb{E}[Y \mid \tau_{1:t-1}, A_t{=}a_t] - \mathbb{E}[Y \mid \tau_{1:t-1}]\Big)}_{(\mathrm{A})}
         + \underbrace{\Big(\mathbb{E}[Y \mid \tau_{1:t}] - \mathbb{E}[Y \mid \tau_{1:t-1}, A_t{=}a_t]\Big)}_{(\mathrm{B})}.
\label{eq:ab-split}
\end{equation}
Term $(\mathrm{B})$ of Eq.~\eqref{eq:ab-split} is the contribution of the
observation: the change in expected outcome from learning $O_t = o_t$, holding the
action fixed. For term $(\mathrm{A})$, the law of total expectation over the
policy's action distribution gives
\[
\mathbb{E}[Y \mid \tau_{1:t-1}]
= \sum_{a} \pi(a \mid \tau_{1:t-1})\, \mathbb{E}[Y \mid \tau_{1:t-1}, A_t{=}a]
= \mathbb{E}_{a\sim\pi}\big[\mathbb{E}[Y \mid \tau_{1:t-1}, A_t{=}a]\big],
\]
where the sum ranges over the support of $\pi(\cdot \mid \tau_{1:t-1})$, so every
conditional expectation appearing in it is well defined. Applying
Proposition~\ref{prop:seqrand} to each term replaces every observational
conditional by its interventional counterpart, giving the action effect relative
to the policy's own baseline,
\begin{equation}
\begin{aligned}
\Delta_t
={}& \underbrace{\Big(
\mathbb{E}[Y \mid \tau_{1:t-1}, do(A_t{=}a_t)]
- \mathbb{E}_{a\sim\pi}\!\left[\mathbb{E}[Y \mid \tau_{1:t-1}, do(A_t{=}a)]\right]
\Big)}_{\substack{\text{action effect}\\ \text{vs.\ policy baseline}}} \\
&\quad
+ \underbrace{\Big(
\mathbb{E}[Y \mid \tau_{1:t}]
- \mathbb{E}[Y \mid \tau_{1:t-1}, A_t{=}a_t]
\Big)}_{\substack{\text{observation}\\ \text{contribution}}}.
\end{aligned}
\label{eq:decomp}
\end{equation}
\qed

\paragraph{Remark.}
Two readings of Eq.~\eqref{eq:decomp} are excluded. The baseline is the policy's
\emph{own} action distribution, so $\Delta_t$ answers how much better or worse a
step was than what this agent would typically have done there, not than the
optimal action --- the latter is a necessity-type contrast requiring intervention
(\S\ref{sec:breaks}). And Eq.~\eqref{eq:telescope} holds for population
quantities, so with finitely many replays per $v_t$ the empirical sum carries
error; we use that error as an internal consistency check rather than reporting it
as exact. The proof uses boundedness of $Y$ only through the existence of the
conditional expectations, so both identities hold verbatim for the recall outcome
$Y_{\mathrm{rec}} \in [0,1]$ and for the per-file indicator $Y_f$, which is what
licenses the outcome choice of \S\ref{sec:practical_estimation} without restating
the theorem.

\subsection{Data-flow reachability is not causal effect}
\label{sec:dataflow}

Data-flow ``inertness'' is sometimes offered as a proxy for causal irrelevance: if
a step's output is never read downstream, the step is taken to have had no effect.
The two notions come apart in both directions, which is why
Table~\ref{tab:consequences} lists this as a distinct failure mode.

\paragraph{Inert but causally potent.}
A \texttt{grep} that returns nothing contributes no content to any subsequent
context window, so it is inert by any data-flow criterion. It may nonetheless
cause the agent to abandon the correct directory, because absence of information
is itself information: the empty result licenses the inference that the target is
elsewhere. Under our estimand this shows up as a large $|\Delta_t|$ at a step with
no outgoing data flow.

\paragraph{Data-carrying but causally inert.}
Conversely, a long file read may inject thousands of tokens into the context while
leaving every subsequent action unchanged. Fixing the prefix and deleting the read
verifies this directly: if the continuation is unchanged across replays, the step's
contribution is zero despite its data-flow footprint being large. The asymmetry
matters because taxonomies that score steps by downstream data flow will
systematically misrank exactly the steps that a causal criterion identifies as
decisive.

\section{Implementation and Experimental Details}
\label{sec:impl-app}

\subsection{Estimation algorithm}
\label{sec:algorithm}

\begin{algorithm}[h]
\caption{SCAE: step contributions with per-step identification}
\label{alg:scae}
\begin{algorithmic}[1]
\Require episode $(q, \text{repo})$, rollout $\tau_{1:T}$, outcome $Y$, estimation
points $\mathcal{P} \subseteq \{1,\dots,T\}$, replays $m$, threshold $\tau_H$
\Ensure $\{\widehat\Delta_t\}_{t \in \mathcal{P}}$, entropy $\{\widehat H_t\}$
\State reset repository to base commit; pin locale, timezone, terminal width, hash seeds
\For{$t \in \mathcal{P}$}
  \State $\tau_{1:t-1} \gets \textsc{TruncateBeforeStep}(\tau, t)$
         \Comment{rewrite rollout log at item granularity}
  \State \textsc{AssertNoDanglingCalls}($\tau_{1:t-1}$)
  \State $\mathcal{A} \gets \emptyset$
  \For{$i = 1$ to $m$}                       \Comment{common random numbers: shared prefix and pinned env}
    \State reset repository; fork agent from $\tau_{1:t-1}$; resume with the
           byte-identical continuation prompt
    \State $\mathcal{A} \gets \mathcal{A} \cup \{\textsc{ActionBucket}(A_t^{(i)})\}$
           \Comment{bucket $=(\mathrm{tool}, \text{target node in } G)$}
  \EndFor
  \State $\widehat H_t \gets \textsc{Entropy}(\mathcal{A})$
  \If{$|\{\mathcal{A}\}| \ge 2$ \textbf{and} $\widehat H_t > \tau_H$}
        \Comment{overlap available: observational branch}
    \State $\widehat v_{t} \gets \frac{1}{m}\sum_i Y^{(i)}$ from replays at $t$;
           $\widehat v_{t-1}$ likewise from replays at $t-1$
    \State $\widehat\Delta_t \gets \widehat v_t - \widehat v_{t-1}$
  \Else                        \Comment{degenerate policy: soft intervention, Asm.~\ref{as:a5}}
    \State draw $a'$ adjacent to $a_t$ in $G$, or an oracle action inspecting a gold file
    \State $o' \gets$ \textbf{execute} $a'$ for real in the reset repository
           \Comment{never a fabricated output}
    \State splice $(a', o')$ into $\tau_{1:t-1}$; resume $m$ times; average to get
           $\widehat{\mathbb{E}}[Y \mid \tau_{1:t-1}, do(A_t{=}a')]$
    \State $\widehat\Delta_t \gets
            \widehat{\mathbb{E}}[Y \mid \tau_{1:t-1}, do(A_t{=}a')]
          - \widehat{\mathbb{E}}[Y \mid \tau_{1:t-1}, do(A_t{=}a_t)]$
  \EndIf
  \State record which branch was taken, so that the reported estimand is labelled
         $\pi$ or $\tilde\pi$
\EndFor
\end{algorithmic}
\end{algorithm}

Two implementation invariants are load-bearing and easy to violate. First, the
continuation prompt must be byte-identical across all arms: it enters the context
and is therefore part of the estimand, so any difference between arms becomes an
undeclared intervention. Second, the spliced observation $o'$ must be the genuine
result of executing $a'$; substituting a plausible-looking output intervenes on the
observation rather than the action, which is a different causal quantity.

\subsection{Experimental settings}
\label{sec:settings}

\paragraph{Agent and model.}
Codex CLI driven over its JSON-RPC app-server with
\texttt{capabilities.experimentalApi} enabled, so that thread paths are exposed
and forking by path is available. Sandbox \texttt{read-only}, approval policy
\texttt{never}; we verify that zero server-side approval requests are received,
since a nonzero count would mean the read-only restriction was not in force.
Backing model \texttt{gpt-5.5-0424-global}. Per-turn timeout $900$\,s.

\paragraph{Environment pinning.}
Locale, timezone, terminal width and hash seeds are fixed; tool outputs are
totally ordered and truncation points fixed. \S\ref{sec:setup} reports the
empirical verification of this pinning, at \ReplayRate{} byte-identical replay
over \ReplayN{} actions.

\paragraph{Estimation points and replays.}
Attribution uses $m = \MReplay{}$ replays per estimation point and $5$ equidistant
estimation points per episode, with the recall outcome of
\S\ref{sec:practical_estimation}. Entropy probes use $4$--$6$ replays per point,
which is why Table~\ref{tab:entropy} reports entropies on a coarse grid: with $4$
replays the attainable nonzero values are $0.56$, $1.04$ and $1.39$ nats. The
entropy threshold is $\tau_H = 0$, so the observational branch is taken exactly
when the replays realize at least two distinct action buckets; this is the most
permissive threshold, and it makes the reported \HighEntFrac{} an upper bound on
how often observational identification is available.

\paragraph{Judge conditions.}
Both judge conditions use the same model and the same rubric, and differ only in
the visible portion of the trace: the full-trace condition sees $\tau_{1:T}$ and
the realized outcome $Y$; the prefix-only condition sees a fixed $60\%$ prefix with
the outcome withheld. The truncation fraction is fixed in advance and does not
depend on any estimate. This matters, and \S\ref{sec:prereg} records why: an
earlier design truncated just before the step selected by the causal estimate,
which would have produced a positive displacement even under the null.

\paragraph{Covariates for $v_0$.}
The \NCov{} pre-execution covariates are: issue description length; count of
explicitly named files; presence of a traceback; error-term density; repository
size; distinct identifier count; and the number of repository files containing
those identifiers, which serves as an ambiguity proxy for $U_{\mathrm{task}}$.
None of these may read the trace, since doing so would leak execution-time
information into a pre-execution predictor.

\subsection{The baseline suite in full}
\label{sec:baselines-app}

Because this paper compares \emph{measurement procedures} rather than agents, each
research question carries its own reference set, and every reference is either a
competing predictor evaluated on the identical points or a null with a closed
form. We list the full inventory here, since \S\ref{sec:setup} only summarizes it.

\paragraph{Next-action prediction (\S\ref{sec:rq1}).}
Nine predictors, all scored on the same decision points with uncovered points
counted as misses rather than dropped (Table~\ref{tab:rq1baselines}): a
repository-frequency marginal; a visit-history frequency model; a
\emph{strengthened} code-dependency-graph predictor that seeds a multi-source
proximity ranking from the entire visited set rather than from the last step alone;
four execution-provenance variants (uniform over previously mentioned paths,
recency-weighted, frequency-weighted, and last-output-only with within-output
position ranking); and two tool-routed candidate-set constructions. We
additionally contrast a factored $P(\mathrm{tool})P(\mathrm{obj})$ model against a
routed one (Table~\ref{tab:joint}). The graph predictor is deliberately
strengthened rather than strawmanned, because the claim of \S\ref{sec:rq1} is that
code structure loses even when given every advantage we could construct.

\paragraph{Graph and reuse structure (\S\ref{sec:rq2}--\S\ref{sec:rq3}).}
A within-episode permutation null, which preserves the multiset of visited objects
and destroys only their order, and random same-size file sets for the region
comparison. The permutation null is the load-bearing choice: it holds the working
region fixed, which is exactly what separates ``the graph describes where the agent
works'' from ``the graph describes how it moves''. Revisit statistics are compared
against closed-form uniform expectations rather than sampled draws.

\paragraph{Execution uncertainty (\S\ref{sec:rq4}).}
Two controls in opposite directions. The superseded Bernoulli entropy
approximation serves as a negative control, and a quantity known to respond
serves as a positive control at $\rho=\RecallUexecRho{}$; without the latter, a
null relating action entropy to replay variance would be uninterpretable, since it
could reflect a dead outcome measure rather than an absent relationship.
Structural probes with no mechanical coupling to the mean (step position, tool
type, instance identity) are used in place of probes that are mechanically
coupled.

\paragraph{Attribution (\S\ref{sec:rq5}).}
Four reference sets. First, \NJudges{} judge models including a code-specialised
one (Table~\ref{tab:crossmodel}). Second, a prefix-only against a full-trace
ablation of the judge's own information set, which is the only comparison here
that needs no external reference. Third, five zero-LLM structural rules
(Table~\ref{tab:judge}), which test whether the judge is reducible to a positional
or provenance heuristic. Fourth, a uniform-random selector reported at its exact
per-episode expectation $1/T_i$ rather than as a sampled draw, so that the chance
level carries no Monte Carlo noise.

\paragraph{Two families we do not implement.}
Back-filled process reward models and ablation-with-restart are absent, and both
leave a gap we state rather than paper over. A trained process reward model might
recover causal structure despite misspecified supervision, which our theoretical
account predicts it will not but our experiments do not test.
Ablation-with-restart, being a different estimand (Table~\ref{tab:consequences}),
would need a separate comparison asking how far the marginal contrast departs from
the history-conditioned one --- a measurable quantity we have not measured.

\subsection{Positivity probe: entropy at each decision point}
\label{sec:entropy-app}

\begin{figure}[h]
\centering
\includegraphics[width=0.85\linewidth]{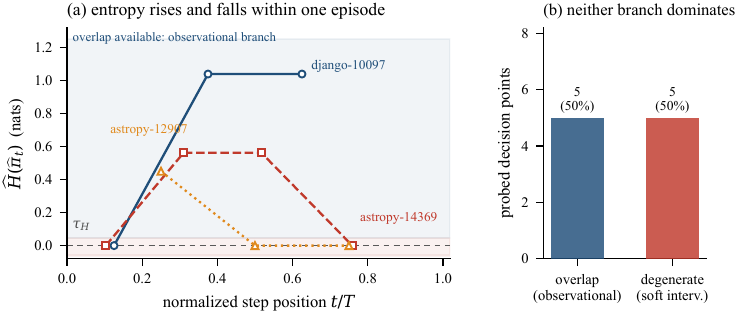}
\caption{\textbf{Overlap is a local property, not a global one}. Each marker is one
probed decision point, positioned by its normalized position in the episode and
coloured by episode. Points on the zero line admit no observational
identification and are routed to the intervention branch of
Eq.~\eqref{eq:hybrid}; points above it are estimated observationally under
Proposition~\ref{prop:seqrand}. \HighEntFrac{} of points lie above zero, and the
rank correlation with position is only \EntSpearman{}, so overlap is not a
monotone function of trajectory depth. We report the sample size plainly:
\NForkPoints{} points establish that overlap varies \emph{within} episodes, which
is all the estimator requires, but not the distribution of that variation, and the
near-zero rank correlation should not be read as evidence of no relationship.}
\label{fig:positivity}
\end{figure}

Identification of Eq.~\eqref{eq:estimand} needs overlap at the step being
estimated, and \S\ref{sec:hybrid} treats this as something to measure rather than
grant. Probing repeated replays from identical prefixes, action entropy is
non-zero at \HighEntFrac{} of \NForkPoints{} probed decision points across
\NForkEpisodes{} episodes, with a rank correlation with normalized position of
\EntSpearman{}. Table~\ref{tab:entropy} gives every probed point and
Figure~\ref{fig:positivity} plots them.

\begin{table}[h]
\centering
\small
\begin{tabular}{@{}llc@{}}
\toprule
\textbf{Episode} & \textbf{Entropy (nats) by normalized position} & \textbf{Pattern} \\
\midrule
\texttt{django-10097} (16 steps) &
$0.12{:}\,0 \rightarrow 0.38{:}\,1.04 \rightarrow 0.62{:}\,1.04$ &
det.--stoch.--stoch. \\
\texttt{astropy-14369} (29 steps) &
$0.10{:}\,0 \rightarrow 0.31{:}\,0.56 \rightarrow 0.52{:}\,0.56 \rightarrow 0.76{:}\,0$ &
det.--stoch.--det. \\
\texttt{astropy-12907} (8 steps) &
$0.25{:}\,0.45 \rightarrow 0.50{:}\,0 \rightarrow 0.75{:}\,0$ &
stoch.--det. \\
\bottomrule
\end{tabular}
\caption{Action entropy at all \NForkPoints{} probed decision points across
\NForkEpisodes{} episodes. Overlap is present at some steps and absent at others
\emph{within the same episode}, and the three episodes realize three different
patterns. This is what forces the per-step branch selection of
Eq.~\eqref{eq:hybrid}: neither a globally observational nor a globally
interventional estimator is correct. Entropies fall on a coarse grid because each
point is probed with $4$--$6$ replays.}
\label{tab:entropy}
\end{table}

\subsection{Statistical conventions}
\label{sec:conventions-app}

Conventions are fixed once and applied throughout, so that no comparison is
reported under a rule chosen after seeing it. Confidence intervals are episode- or
instance-level bootstrap. Nulls are within-episode permutation or closed-form
expectation, and never sampled where a closed form exists. Ranking metrics use the
closed-form expectation over random tie-breaking, since count-based scorers tie
heavily and a single arbitrary tie-break would be an undeclared design choice.
Predictors with different candidate-set sizes are compared on a \emph{common
denominator}, with points whose realized action falls outside the candidate set
scored as misses rather than dropped --- otherwise a predictor could buy accuracy
by abstaining. Multiple comparisons use BH-FDR at $q=0.10$, and we label
exploratory versus confirmatory findings accordingly. Our Wilcoxon implementation
declines to compute a $p$-value below $n=6$ rather than reporting an unreliable
one.

\subsection{Choice of outcome variable}
\label{sec:outcome-app}

\S\ref{sec:practical_estimation} adopts recall over the gold file set. The choice
was made by screening six candidate outcomes on intervention sensitivity, measured
as the paired effect divided by its standard deviation: recall attains
\RecallSens{}, against \FOneSens{} for $F_1$ and \YSens{} for binary success. The
second criterion is headroom: \PctMoved{} of instances can move under recall
against only \PctMovedY{} under the binary outcome, whose failures cannot fall
further.

The mechanism behind those numbers is worth stating, because it is the reason a
formally adequate outcome can still be empirically useless. With $Y$ binary at the
task level, single-target instances are solved almost always, so
$\Delta_t \equiv 0$ and attribution says nothing. Refining to the per-file
indicator $Y_f$ localizes the question to which step decided whether $f$ was
found, but degenerates whenever $f$ is unreachable under every prefix. Recall
retains signal from the remaining targets when one is hopeless, which is what
reduces the degenerate rate to the \DegenRate{} reported in \S\ref{sec:rq4}.
Theorem~\ref{thm:additive} requires only boundedness, so this substitution changes
no formal result (\S\ref{sec:proofs}).

\section{Additional Results}
\label{sec:results-app}

\subsection{Per-repository localization}
\label{sec:byrepo}

\begin{table}[h]
\centering
\small
\begin{tabular}{@{}lcc@{}}
\toprule
\textbf{Repository} & $n$ & \textbf{Localization success} \\
\midrule
\texttt{django}       & $230$ & $93.5\%$ \\
\texttt{sympy}        & $75$  & $94.7\%$ \\
\texttt{sphinx-doc}   & $44$  & $88.6\%$ \\
\texttt{matplotlib}   & $34$  & $91.2\%$ \\
\texttt{scikit-learn} & $32$  & $100.0\%$ \\
\texttt{astropy}      & $22$  & $100.0\%$ \\
\texttt{pydata}       & $22$  & $90.9\%$ \\
\texttt{pytest-dev}   & $19$  & $100.0\%$ \\
\texttt{pylint-dev}   & $10$  & $90.0\%$ \\
\texttt{psf}          & $8$   & $100.0\%$ \\
\texttt{mwaskom}      & $2$   & $100.0\%$ \\
\texttt{pallets}      & $1$   & $100.0\%$ \\
\midrule
Total & $\NInstances{}$ & \SolveRate{} \\
\bottomrule
\end{tabular}
\caption{Localization success by repository under the criterion
$F^\star \subseteq \hat F$. The corpus is dominated by \texttt{django} at $46\%$
of instances, which is why leave-one-repository-out is the evaluation protocol for
$v_0$ (\S\ref{sec:v0-app}) and why repository-conditional calibration is used for
conformal sets (\S\ref{sec:conformal}). Higher is better.}
\label{tab:byrepo}
\end{table}

The high aggregate success rate is itself a constraint on what the corpus can
support. It is why the binary outcome has almost no headroom
(\S\ref{sec:outcome-app}), why attribution is restricted to the multi-target
subset, and why a constant predictor is nearly calibrated --- which in turn is why
\S\ref{sec:v0-app} reports selective-risk quantities alongside ECE rather than ECE
alone.

\subsection{Additional measurements behind the action-level result}
\label{sec:action-app}

\S\ref{sec:rq1} reports the headline predictor comparison. Four supporting
measurements are recorded here.

\paragraph{The provenance set, not the repository, is the action space.}
Agents mention \MentionedPerEp{} distinct files per episode and actually visit
\VisitedPerEp{}, a conversion rate of \MentionToVisit{}. This is the mechanism
behind Table~\ref{tab:rq1baselines}: the candidate set an agent draws from is two
orders of magnitude smaller than the repository and is constructed by the agent's
own earlier outputs. It also explains why the strengthened graph predictor carries
a median effective candidate count in the thousands while the provenance predictor
carries \ProvEff{} --- the graph ranks the repository, whereas provenance ranks
what has been seen.

\paragraph{Using the full history rather than the last output is not free.}
Reading the entire observation history rather than only the most recent output adds
$+0.011$ (CI $[+0.005,+0.019]$) at top-3 while inflating effective candidates from
\ProvEff{} to $18.8$. The last-output variant is therefore the one we report as
headline, since it is both sharper and cheaper; the comparison is what rules out
the reading that provenance works merely by accumulating a large candidate pool.

\paragraph{Sample-size expansion from dropping the code graph.}
Provenance requires only string normalization, because agent commands and shell
outputs draw paths from the same vocabulary. Dropping the code graph therefore
raises the evaluable sample from \NPtsGraph{} decision points to \NPts{} points
over \NEpEval{} episodes, out of \NEpAll{} episodes usable without a graph. We
verified that this expansion does not shift conclusions: on the overlapping
graph-built subset the headline metric differs by under $20\%$, and the residual
gap is explained by object-type composition rather than by normalization. The
strengthened graph predictor's top-3 on first-visit steps is \GraphTopThree{}
(CI \GraphTopThreeCI{}).

\paragraph{Tool and object type are dependent, and the routed model stays
calibrated.}
The mutual information between tool identity and object type is \ToolObjMI{} nats
against a permutation null of \ToolObjMINull{} ($p=\ToolObjMIP{}$), which is what
makes the factored $P(\mathrm{tool})P(\mathrm{obj})$ model misspecified rather than
merely suboptimal. Routing does not buy accuracy at the cost of calibration: the
routed model retains ECE $=\RoutedECE{}$ (Table~\ref{tab:joint}).

\subsection{Additional measurements behind the graph and reuse results}
\label{sec:graph-app}

\paragraph{Raw rates behind the permutation lifts.}
The one-hop lift of \OneHopLift{} reported in \S\ref{sec:rq2} is the ratio of an
observed adjacent-step one-hop rate of \OneHopObs{} to a within-episode permutation
expectation of \OneHopNull{}, with immediate self-repeats removed. We report the raw
pair because the lift alone can be read as evidence that adjacency is rare, which it
is not: roughly $28\%$ of consecutive steps \emph{are} one hop apart, and the point
is that a null which preserves the visited multiset predicts the same rate. This is
the sense in which the graph describes the region rather than the movement.

\paragraph{Leakage control in the gold-ranking measurement.}
\GoldVisitedShare{} of gold files have already been visited by the time the ranking
is computed, so leaving them in the candidate set would restate the agent's own
history rather than predict anything. The leakage-controlled figures of
\S\ref{sec:rq2} are computed on unvisited gold files only, which is why they rest on
$n=39$ (unseen) and $n=54$ (prefix-seeded) instances --- small samples that we
prefer to a larger but contaminated one.

\paragraph{Sample for the reuse analysis.}
The reuse statistics of \S\ref{sec:rq3} are computed over \NEpReuse{} episodes.
The permutation null for the reuse \emph{rate} has standard deviation exactly zero
in our data, which is the formal sense in which that rate carries no information:
the number of first occurrences among positions $2..n$ is exactly
$\#\text{distinct}-1$, so the reuse rate is a function of the visited multiset and
is invariant to order.

\subsection{The query channel and a measured parser defect}
\label{sec:query-app}

\S\ref{sec:rq1} reports that steps taking a \emph{novel} free-text search query
are predicted at top-3 $=\TopThreeQuery{}$, and treats that as the binding
constraint on next-action prediction. The number deserves scrutiny, because our
command parser recovers the search pattern via \texttt{shlex}, which fails on
\ShlexFailRate{} of command segments --- an unbalanced quote such as
\verb|grep "a\|b"| --- and then falls back to splitting on whitespace, retaining
the quote character and truncating. \QueryDamaged{} of query objects are affected,
and \QueryFragDef{} of them collapse onto the single fragment \texttt{q:"def}. We
therefore re-extracted the patterns with a parser that does not depend on
\texttt{shlex} succeeding, and re-ran the channel over \NQueryPts{} query steps.

The repair moves the number \emph{down}, not up: those frequent fragments were a
spurious repetition mode, and removing them costs \QueryParseDelta{} in top-3
(CI $[-0.049,-0.022]$), taking the same model from \QueryTopThreeOld{} to
\QueryTopThreeRepaired{}. Improving the model then recovers ground. Weighting
history by frequency rather than recency helps marginally, and adding identifiers
mined from the issue text helps substantially (\QueryProblemGain{},
CI $[+0.029,+0.054]$, with coverage rising from $0.060$ to \QueryCovBest{}), while
adding symbol names harvested from prior tool \emph{outputs} actively hurts
(\QuerySymbolGain{}). Our best configuration reaches top-3
$=\QueryTopThreeBest{}$ at coverage \QueryCovBest{}, and the net change from the
original pipeline is not significant ($+0.002$, CI $[-0.016,+0.019]$).

Three consequences follow. The conclusion that queries bound next-action
prediction \emph{survives}, and is now the outcome of a deliberate attempt rather
than of a weak baseline. The defect biases in the conservative direction for that
claim --- it made queries look more predictable than they are --- so it does not
threaten it. And there is a positive mechanism worth recording, which contrasts
instructively with the path channel of \S\ref{sec:rq1}: an agent's \emph{paths}
come from what it just read in an output, whereas its \emph{query terms} come from
the issue text, and importing symbol names from outputs into the query model only
dilutes it. We report the defect rather than silently fixing it because its
direction is what licenses the claim to stand.

\subsection{A binary criterion that manufactured an exception}
\label{sec:binary-app}

An earlier pass over the data of \S\ref{sec:rq4} thresholded the per-point test at
$p<0.05$ and found that high-entropy steps carried a nominally higher rate of
significant effects, which would have contradicted that section's null. It does
not replicate. The stratified difference is $+0.056$ with CI \EffectDiffCI{}, and
replacing the binary indicator with either $|\Delta_t|$ ($\rho=\HAbsDeltaRho{}$)
or $-\log_{10}p$ leaves no relationship at all. The diagnostic is the threshold
sweep: as $\alpha$ is relaxed and the event count grows from $8$ to $24$, the
ratio decays monotonically from \RatioAlphaFive{} to \RatioAlphaTwenty{} --- the
signature of a small-sample artefact rather than of an effect. We record it
because dichotomising a continuous quantity produces clean-looking strata at small
$n$, and the continuous version plus a threshold sweep is what distinguishes the
two.

\subsection{Uncertainty components and their estimators}
\label{sec:uq-table}

\begin{table}[h]
\centering
\small
\begin{tabular}{@{}llll@{}}
\toprule
\textbf{Component} & \textbf{Type} & \textbf{Estimated by} & \textbf{Implied action} \\
\midrule
$U_{\mathrm{task}}$ & aleatoric, irreducible &
residual; graph ambiguity covariates & ask the user to disambiguate \\
$U_{\mathrm{policy}}$ & aleatoric, reducible &
between-rollout variance at fixed $q$ & resample and aggregate \\
$U_{\mathrm{model}}$ & epistemic &
posterior variance through the link & abstain, escalate, collect data \\
\bottomrule
\end{tabular}
\caption{The three components of Eq.~\eqref{eq:decomp3}. Each answers a different
question, so conflating them discards the decision-relevant content: a system that
reports only total uncertainty cannot distinguish ``the request is ambiguous, ask
the user'' from ``the model is out of its depth, escalate''.}
\label{tab:uq}
\end{table}

\begin{figure}[h]
\centering
\includegraphics[width=\linewidth]{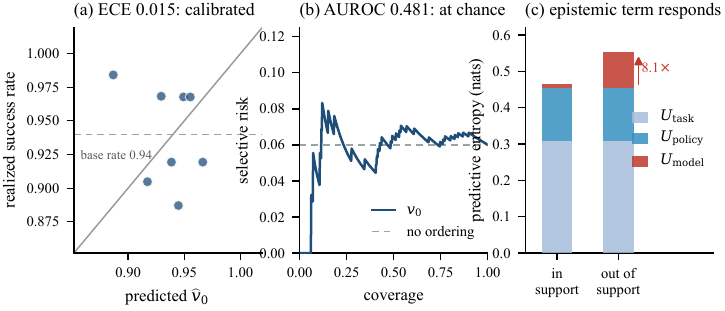}
\caption{\textbf{The uncertainty decomposition, validated synthetically and
unvalidated empirically}. The panels report the forecasting behaviour of the left
endpoint $v_0$ on the real corpus ($n=\NInstances{}$, leave-one-repository-out
AUROC \AUROC{}, ECE \ECE{}) alongside the synthetic self-test in which the
generating process is known. On the real corpus the risk--coverage curve is
essentially flat at the base error rate, which is the signature of a calibrated
but uninformative predictor and is why we report AURC and AUROC rather than ECE
alone. On the synthetic process the decomposition behaves as specified, with
epistemic uncertainty inflating $\StRatio{}\times$ outside the training support.
The contrast is the point: the machinery works where the truth is known and does
not deliver a usable forecast where it is not.}
\label{fig:uq}
\end{figure}

\paragraph{Status on our data.}
We state this plainly, because it is weaker than the derivation suggests. Only
$U_{\mathrm{policy}}$ is estimated directly, at \UPolicy{} from the \NUPolicy{}
instances carrying $K=\KRep{}$ repeated rollouts. The three-way separation is
validated on a \emph{synthetic} generating process with known truth, not on these
traces: parameter recovery reaches correlation \StWCorr{}, mean absolute
probability error is \StProbMAE{}, and predictive entropy \StH{} splits as
$\StUTask{}/\StUPolicy{}/\StUModel{}$ across the three components, with
$U_{\mathrm{model}}$ inflating from \StUModel{} in support to \StUModelOOD{} out
of support, a factor of \StRatio{}. Figure~\ref{fig:uq} shows the self-test
together with the forecasting result it does not rescue. The identifiability of
the three terms on real traces is therefore \emph{not} established here, and the
falsifiable prediction that additional sampling cannot help where
$U_{\mathrm{task}}$ dominates is \emph{not} tested, because it presupposes the
calibrated $v_0$ that \S\ref{sec:v0-app} reports as absent.
Table~\ref{tab:uq} is a specification with a synthetic self-test attached, and
should be read as such.

\subsection{Estimating the left endpoint \texorpdfstring{$v_0$}{v0}}
\label{sec:v0-app}

Our corpus has strong group structure by repository and a high base rate, which is
a small-data regime. We therefore fit $v_0$ as a Bayesian logistic model with a
Gaussian prior $w \sim \mathcal{N}(0, I)$ (precision $\lambda = 1$), taking the MAP
by IRLS and the posterior covariance $\Sigma$ by Laplace approximation, then
propagating parameter uncertainty by the probit approximation to obtain
$U_{\mathrm{model}} \approx p(1-p)\, x^\top \Sigma x$. No hyperparameter search was
performed. Evaluation is leave-one-repository-out over \NRepos{} repositories. The
claims at issue concern calibration and identifiability rather than predictive
ceiling, so a small interpretable model with a tractable posterior serves them
better than a tuned ensemble whose uncertainty is heuristic.

The measured outcome is negative. Leave-one-repository-out AUROC is \AUROC{} over
\NCov{} covariates, which is at chance against a base rate of \BaseRate{}, with
ECE and Brier score as defined in Eq.~\eqref{eq:calib} reaching \ECE{} and
\Brier{}, and area under the risk--coverage curve \AURC{}. The low ECE is not reassuring: a high base rate makes calibration easy to
satisfy by a nearly constant predictor, and the flat risk--coverage curve confirms
that the predictor supplies no usable ordering. We therefore report this as a
negative result rather than as a calibrated forecaster, and it is the reason the
falsifiable prediction attached to Eq.~\eqref{eq:decomp3} is untested
(\S\ref{sec:uq-table}). The predictor may read only $q$ and $G$, never the trace,
since doing so would leak execution-time information.

\subsection{Set-valued prediction and exchangeability}
\label{sec:conformal}

Localization is naturally set-valued, so alongside the scalar $v_0$ we report
conformal sets $\mathcal{C}(q)$ satisfying
$P(F^\star \subseteq \mathcal{C}(q)) \ge 1-\alpha$, together with average set
size, since a guarantee attained by returning the whole repository is vacuous.
Exchangeability requires care rather than assertion in this setting: traces are
sequentially dependent, the observations an agent collects are self-selected by its
own earlier actions, and repositories are heavily imbalanced
(Table~\ref{tab:byrepo}). We therefore use split conformal with
repository-conditional (Mondrian) calibration rather than pooled calibration,
report group-conditional coverage gaps rather than marginal coverage alone, and
apply covariate-shift weighting for cross-repository transfer. Marginal coverage
that is correct on average while failing badly on the smaller repositories is the
failure mode this guards against, and pooled calibration on a corpus that is
$46\%$ \texttt{django} would conceal it.

\subsection{Additional measurements behind the attribution results}
\label{sec:judge-app}

\S\ref{sec:rq5} reports the displacement and enrichment results. The supporting
quantities are recorded here, since several of them are what fix the interpretation.

\paragraph{Absolute positions, not just the displacement.}
The collider contrast moves the verdict from a mean normalized position of
\PrefixPosMean{} under the prefix-only condition to \JudgePosMean{} under the
full-trace condition, with a median displacement of \ColliderMedian{}. The
absolute positions matter for reading the effect: the prefix-only judge is not
merely earlier, it sits in the first quarter of the trajectory, while the
full-trace judge sits in the last quarter. For comparison, the intervention-based
selector has mean position \CausalPosMean{}. Reporting only the difference would
leave open the possibility that both conditions cluster in the same region.

\paragraph{Enrichment interval and the comparable subset.}
The full-trace judge places \JudgeGold{} (CI \JudgeGoldCI{}) of its top-1 verdicts
on gold-provenance steps. The restriction to steps that are directly comparable
between the judge and the intervention estimate rests on $n=\NFair{}$ episodes,
because only \JudgeInEst{} of judge verdicts fall inside the estimator's window
set at all. The two baselines are therefore computed separately over their own
selection spaces, and the comparable-subset contrast is reported alongside the
unrestricted one rather than instead of it.

\paragraph{Cross-model budget and per-judge enrichment.}
The replication uses one additional general-purpose model (\XModel{}) and one
code-specialised model, at \XCalls{} API calls per judge, with the prompt held
byte-identical and the trace set fixed. Displacement is positive in all
\JudgesPositive{} of \NJudges{} judges. Per-judge enrichment on gold-provenance
steps is \EnrichOrigCommon{} for the original judge, \EnrichQwen{} for the second
general model and \EnrichCoder{} for the code-specialised one, against the
common-episode baseline of \EnrichBaseCommon{}
(Table~\ref{tab:crossmodel}). We note but do not claim the pattern that the
code-specialised model behaves like the original rather than like its own
general-purpose sibling: \NJudges{} judges cannot separate that explanation from
model scale, training data or rubric sensitivity, so we record it as a hypothesis.

\paragraph{Attribution subsample composition.}
Attribution is computed over \NAttr{} episodes drawn from \NAttrRepos{}
repositories, of which \NAttrFail{} are failing episodes. The
mean replay variance across estimation segments is \UexecMean{}, and the mean
normalized position of the high-entropy steps identified by the positivity probe is
\HighEntPos{} --- close to the middle of the trajectory, which is the sense in
which overlap is not concentrated early.

\subsection{Why \texorpdfstring{$D_1$}{D1} is retired}
\label{sec:metrics-app}

$D_1$ appears in Eq.~\eqref{eq:d1d3} for completeness and because we
pre-registered it, but it is not a valid test and we report it nowhere as
evidence. A disagreement rate is informative only against a reference whose own
selection is reliable. Ours is not: \S\ref{sec:rq4} shows the identified top-1 does
not survive FDR correction (\StepSigBH{} of \NSeg{} effects), so it behaves as a
draw over the $|\mathcal{P}_i|\!\approx\!5$ estimation points while a judge selects
among $T_i\!\approx\!20$ steps. Two selectors drawing from spaces of such different
granularity disagree with probability close to one \emph{under the null}, which is
what we observe (\DisagreeRate{}). The criterion was ill-posed rather than merely
conservative, and we flag this rather than quietly dropping the metric, since
reporting $D_1$ as a positive finding is a mistake another group could repeat.

The same defect propagates to the three statistics of Eq.~\eqref{eq:rz}. $R$, $Z$
and $\bar\Delta$ are all computed \emph{against} $\widehat\Delta$, so each inherits
its noise, and none can adjudicate between ``the judge applies a different but
defensible criterion'' and ``the judge extracts no causal signal''. We retain them
in \S\ref{sec:metrics} as definitions, note that only \NNondegen{} episodes are
non-degenerate and \ZeroPointFrac{} of estimation points have $\widehat\Delta$
exactly zero, and report them nowhere as evidence. The comparisons that do
adjudicate need no causal reference: the provenance-class enrichment and the
information-set isolation contrast, both in \S\ref{sec:rq5}.

\subsection{Comparison with annotation-based process evaluation}
\label{sec:procctrl-compare}

Process-level evaluations of coding agents \citep{procctrl2027} and SCAE target
different estimands, and the difference determines what each can license.

\paragraph{Estimands.}
That line of work estimates $p_j = P(y_j = 1 \mid x_j, c)$, the probability that a
human annotator would mark defect $j$ given trajectory evidence $x_j$ and context
$c$. This is a measurement-theoretic quantity: it is about agreement between a
detector and a rater, and it is identified from logged trajectories without any
causal assumption, since nothing is intervened upon. SCAE estimates
$\mathbb{E}[Y \mid \tau_{1:t-1}, do(A_t{=}a')]$, the effect on the terminal
outcome of substituting one step. This requires the logging policy's randomization
to be preserved (Proposition~\ref{prop:seqrand}) and is \emph{not} identified under
context compaction (\S\ref{sec:breaks}).

\paragraph{Calibration targets.}
The two therefore calibrate against different things. There, calibration means
detector-versus-annotator agreement, and a well-calibrated detector is one whose
confidence matches the rate at which raters agree. Here, calibration means
predicted-versus-realized outcome, and a well-calibrated $v_0$ is one whose
predicted success probability matches the realized success frequency
(\S\ref{sec:v0-app}). Neither target substitutes for the other: a detector
perfectly aligned with annotators tells us nothing about whether the step it flags
changed the outcome, and a perfectly calibrated outcome predictor tells us nothing
about whether humans would call any particular step defective.

\paragraph{Why this is not a criticism.}
That work does not claim causal identification, so the comparison is a delineation
of scope rather than an objection. It is, however, the reason we do not adopt
annotation agreement as a validation target for attribution: a step can be
unanimously judged defective and have zero causal contribution, and a step can
look unremarkable to every annotator and be decisive, as \S\ref{sec:dataflow}
illustrates.

\section{Reproducibility, Pre-registration, and Falsifiability}
\label{sec:repro-app}

\subsection{Reproduction commands and sensitivity analyses}
\label{sec:sensitivity-app}

\paragraph{Commands.}
Every reported number and every data figure is regenerated by a CPU-only script
that fixes its random seed and writes its output to JSON. The uncertainty
self-test of \S\ref{sec:uq-table} is reproduced by
\texttt{python3 -m pilot.analyze.uq\_selftest}; the entropy summary of
\S\ref{sec:entropy-app} by \texttt{python3 -m pilot.analyze.entropy\_scan}; the
localization and forecasting tables by
\texttt{python3 -m pilot.analyze.full\_report}; and the data figures by
\texttt{python3 paper/figs/make\_figures\_rq.py} and
\texttt{python3 paper/figs/make\_figures.py}. The figure scripts read only the
recorded rollout outputs, so plotted and tabulated values cannot drift apart. A
further script, \texttt{python3 paper/analysis/verify\_macros.py}, checks every
numeric macro in the manuscript against the analysis output that produced it, and
\texttt{python3 paper/analysis/check\_tex.py} verifies that every reference,
label, macro, citation and figure path resolves.

\paragraph{Outcome definition.}
\S\ref{sec:setup} reports the primary criterion $F^\star \subseteq \hat F$
alongside exact set equality. The gap between \SolveRate{} and \ExactRate{} is
accounted for entirely by the agent naming test files that gold patches exclude: a
filtered variant that discards test paths before comparison coincides with the
primary criterion to within rounding. We therefore treat exact equality as a
diagnostic of the agent's naming behaviour rather than as a metric of localization
ability.

\paragraph{Effect of the leakage reset.}
Every number is measured after a hard reset of the repository snapshot to its base
commit. This is a silent confound rather than a detail: the same pipeline run
against the distributed snapshots as shipped would measure an agent that can read
the answer out of its own working tree, and would report inflated localization
accuracy with no visible symptom.

\paragraph{Measurement error in regression diagnostics.}
Estimated contributions carry Monte Carlo error, which attenuates regression
coefficients toward zero. For the hypothesis that data-flow reachability fails to
predict causal effect (\S\ref{sec:dataflow}), this bias is \emph{favourable} to us,
so we state its direction explicitly. A correction such as SIMEX, or a larger $m$,
is required before that null is asserted rather than merely observed.

\paragraph{Discrete estimation points.}
Because $v_t$ is estimated at equidistant points rather than at every step,
reported contributions are segment effects and their sum does not satisfy
Eq.~\eqref{eq:telescope} exactly. We therefore present no additivity check as
evidence; a genuine test requires contiguous windows. Equidistant rather than
contiguous-tail points was a deliberate choice, because a window concentrated late
in the trace would bias the position comparison on which \S\ref{sec:rq5} depends.

\subsection{Pre-registration record}
\label{sec:prereg}

The directions, intervals and falsification thresholds in
Table~\ref{tab:prereg} were fixed before any attribution experiment was run. We
report the record in full, including the four entries that did not hold and the one
criterion we retire rather than bank.

\begin{table}[h]
\centering
\small
\begin{tabular}{@{}p{0.19\textwidth}p{0.28\textwidth}p{0.45\textwidth}@{}}
\toprule
\textbf{Quantity} & \textbf{Pre-registered} & \textbf{Outcome} \\
\midrule
$D_1$ top-1 disagreement &
$[40\%,75\%]$, point $55\%$; falsified if $<20\%$ &
\textbf{criterion retired, not passed}: observed \DisagreeRate{}
(\DisagreeCount{}$/$\NAttr{}), CI \DisagreeCI{}. We initially recorded this as
``direction confirmed''; that reading is withdrawn, because the null itself sits
near \DisagreeRate{} once the reference is known to be noise
(\S\ref{sec:metrics-app}) \\
$D_3$ displacement &
$[+0.25,+0.40]$; falsified if $\le 0$ &
\textbf{above interval}: \Displacement{} over $n=\NDisp{}$, $z=\DispZ{}$,
$p<0.001$ \\
Collider isolation &
$[+0.15,+0.30]$; sign is the substantive claim &
\textbf{above interval}: \ColliderIso{} over $n=\NColl{}$, $z=\CollZ{}$,
$p<0.001$. Sign confirmed; magnitude implies descendant conditioning accounts for
most of the displacement \\
Wilcoxon $p$ &
expected to be uncomputable at small $n$ &
superseded: the full attribution sweep makes it computable, and both signed-rank
tests reject at $p<0.001$ \\
Degenerate $\Delta\equiv 0$ rate &
$30$--$40\%$ &
\textbf{better than predicted}: \DegenRate{} (\DegenCount{}$/$\NAttr{}). An early
two-episode reading suggested $50\%$; the full sweep does not support it \\
$U_{\mathrm{policy}}$ estimable &
requires $K\ge 2$; not available from the main sweep &
met on the $K{=}\KRep{}$ subset: $U_{\mathrm{policy}} = \UPolicy{}$ over
\NUPolicy{} instances \\
Entropy vs.\ position &
monotone decay, $\rho \le -0.2$ (``early-high, late-low'') &
\textbf{not supported}: $\rho = \EntSpearman{}$ over \NForkPoints{} points;
reported as non-monotone with episode-dependent inflection \\
$v_0$ LORO AUROC &
$\ge 0.60$ for the forecasting endpoint to be useful &
\textbf{not met}: \AUROC{}, at chance; reported as a negative result
(\S\ref{sec:v0-app}) \\
$U_{\mathrm{model}}$ inflation &
$>2\times$ out of training support &
met: \StUModel{} $\rightarrow$ \StUModelOOD{}, a factor of \StRatio{} \\
Replay consistency &
$\ge 0.99$ byte-identical &
met: \ReplayRate{} over \ReplayN{} actions, zero mismatches \\
\bottomrule
\end{tabular}
\caption{The pre-registration record and its outcomes. Three directional
predictions held at magnitudes above the intervals we fixed in advance, which we
report as the intervals having been too conservative rather than as successes.
Four entries did not hold as stated: the monotone entropy decay is contradicted,
the forecasting endpoint failed its usefulness threshold, the degenerate-outcome
rate came in below the predicted range, and the $D_1$ criterion was ill-posed.}
\label{tab:prereg}
\end{table}

\paragraph{What this record does not cover.}
The decisive-step recovery rate $R$, the mean blamed contribution $\bar\Delta$ and
the null-blame rate $Z$ of Eq.~\eqref{eq:rz} are absent from the table because
they were not pre-registered. We introduced them after observing the displacement,
to distinguish ``the judge applies a different but defensible criterion'' from
``the judge extracts no causal signal''. We now withdraw that use of them, for the
reason given in \S\ref{sec:metrics-app}. Both comparisons that do adjudicate ---
provenance-class enrichment and the information-set isolation contrast --- were
reachable from the same recorded rollouts, so the correction cost no additional
agent execution.

\paragraph{The defect the record caught.}
Our initial design truncated the prefix-only judge just before the step selected by
the causal estimate. Because a judge cannot select a step it cannot see, this caps
the prefix-only condition's reachable positions below the comparison point, and
would have produced a positive displacement even if conditioning on descendants
had no effect at all. Writing the falsification criterion forced us to ask what
would happen under the null, which is what surfaced the defect. The corrected
design uses a truncation fraction fixed in advance and independent of any
estimate (\S\ref{sec:settings}).

\subsection{What would falsify the framework}
\label{sec:discussion}

\S\ref{sec:limitations} states the limitations that bound how our results should be
read. This subsection states the sharper thing: the observations that would
undercut the central argument rather than merely bound it.

Three would undercut the framework itself. If prefix replay failed to reproduce
agent behaviour --- that is, if forking from a reconstructed prefix systematically
diverged from the original continuation --- then Assumption~\ref{as:a1} would be
false in practice and the identification result would not apply to real harnesses.
If the isolation contrast of \S\ref{sec:rq5} produced no displacement at scale, the
collider-bias explanation of judge behaviour would be wrong even though the
structural argument stands; this one has now been run, and the displacement is
\ColliderIso{} with a sign test at $p=\ColliderSignP{}$. And if action entropy were
zero at essentially all decision points, the observational branch of
Eq.~\eqref{eq:hybrid} would never apply and the framework would reduce to
intervention-based ablation with respect to a perturbed policy.

Three further falsifiers attach to the specific empirical claims. If a code-graph
predictor in \emph{any} form matched execution provenance on first-visit steps,
\S\ref{sec:rq1}--\S\ref{sec:rq2} would collapse; we strengthened the graph
predictor deliberately, seeding a multi-source proximity ranking from the whole
visited set, and it still loses, but a better graph formulation is the obvious line
of attack. If the residualised instance-level ICC of \S\ref{sec:rq4} fell back to
its permutation null, then replay variance would carry no detectable
non-mechanical structure at $m=\MReplay{}$, and our null result about structural
entropy would revert to a statement about power rather than about the world --- which
is why we report that ICC before, not after, the null it licenses. And if the
judge's placement on gold-provenance steps regressed to its \GoldStepShare{}
baseline under a different judge or rubric, the characterisation of what such
judges measure would not generalise beyond the ones we tested.

\end{document}